\documentclass[onecolumn]{cohere}
\usepackage{bbm}

\usepackage{latexsym}
\usepackage{caption}
\usepackage{subcaption}
\usepackage{graphicx}
\usepackage{tikz}
\usepackage{booktabs}
\usepackage{multirow}
\usepackage{float}
\usepackage{url}
\usepackage{amsmath}
\usepackage{cleveref}
\usepackage{hyperref}
\usepackage{enumitem}
\usepackage{xcolor}
\usepackage{tabularx}

\definecolor{purp}{HTML}{791f87}

\newcommand{\smallsim}{{\small $\sim$}}

\definecolor{red}{rgb}{0.84,0.153,0.153}
\definecolor{green}{rgb}{0.172,0.627,0.172}

\title{Critical Learning Periods: Leveraging Early Training Dynamics for Efficient Data Pruning}

\author{
    name={Everlyn Asiko Chimoto$^{1,2,3}$ \quad Jay Gala$^{1,5}$ \quad Orevaoghene Ahia$^{1,6}$
    
    Julia Kreutzer$^{7}$ \quad Bruce A. Bassett$^{2,4}$ \quad Sara Hooker$^{7}$ \vspace{0.5em}},
    affiliation={$^{1}$Cohere For AI Community \quad $^{2}$University of Cape Town, South Africa \\
    
    $^{3}$African Institute for Mathematical Sciences \quad $^{4}$South African Astronomical Observatory \\
    
    $^{5}$Mohamed bin Zayed University of Artificial Intelligence \\ [1ex]
    $^{6}$University of Washington $^{7}$Cohere For AI}
}

\date{\today}

\abstract{
Neural Machine Translation models are extremely data and compute-hungry. However, not all data points contribute equally to model training and generalization. Data pruning to remove the low-value data points has the benefit of drastically reducing the compute budget without a significant drop in model performance. In this paper, we propose a new data pruning technique: \emph{Checkpoints Across Time} (\emph{CAT}), that leverages early model training dynamics to identify the most relevant data points for model performance. We benchmark \emph{CAT} against several data pruning techniques including COMET-QE, LASER and LaBSE. We find that \emph{CAT} outperforms the benchmarks on Indo-European languages on multiple test sets. When applied to English-German, English-French and English-Swahili translation tasks, \emph{CAT} achieves comparable performance to using the full dataset, while pruning up to 50\% of training data.  We inspect the data points that \emph{CAT} selects and find that it tends to favor longer sentences and sentences with unique or rare words.
}

\begin{document}

\section{Introduction}\label{sec:intro}

In recent years, there has been significant improvement in the quality of Neural Machine Translation (NMT) models \citep{johnson-etal-2017-googles,Arivazhagan2019MassivelyMN,liu-etal-2020-multilingual-denoising,m2m100,reid-etal-2021-afromt,ramesh-etal-2022-samanantar,nllbteam2022,Bapna2022BuildingMT,indictrans2}. The success of these models is primarily due to many factors including pretraining data size, compute abundance and the ever-increasing model size. Despite the gains due to growing datasets, large dataset sizes have also posed significant hurdles for maintaining data quality. Significant portions of web-scraped data used for language model pretraining have been shown to be of low quality, machine-generated spam, pornographic content~\citep{kreutzer-etal-2022-quality}. Given the ever-growing size of parallel corpora, it often becomes laborious for humans to assess the quality at scale \citep{freitag-etal-2021-experts,Agarwal_2022_CVPR,longpre2023data}. Quality of pretraining data for large NMT models often faces amplified quality issues because of extensive bitext mining approaches to automatically extracting translation pairs from either monolingual corpora \citep{guo-etal-2018-effective,schwenk-etal-2021-wikimatrix,schwenk-etal-2021-ccmatrix,nllbteam2022,ramesh-etal-2022-samanantar,vegi-etal-2022-webcrawl,indictrans2} or document-aligned corpora \citep{banon-etal-2020-paracrawl,steingrimsson-etal-2021-effective,ramesh-etal-2022-samanantar,indictrans2}. Bitext mining approaches, although far more cost-effective than human curation, have been shown to introduce significant noise into datasets by relying on sub-optimal sentence embedding models \citep{thompson-koehn-2019-vecalign,feng-etal-2022-language,heffernan-etal-2022-bitext} to match sub-optimal translation pairs \citep{kreutzer-etal-2022-quality}.

Recent work has shown that quality matters. Large neural models trained on high-quality data outperform models trained on noisy data \citep{khayrallah-koehn-2018-impact,arora2021Studying,abdulmumin-etal-2022-separating,indictrans2}  or match performance using far fewer data points \citep{sorscher2023neural, siddiqui2022metadata,DBLP:journals/corr/abs-2201-05700,10.1007/978-3-030-05677-3_10,xu2023better}. In this work, we ask if we can arrive at a rigorous estimator of data quality through \textit{data pruning}. Our goal is to focus on metrics which are scalable across large datasets, and which identify a subset that preserves performance.

We leverage distinctive periods of training to indicate examples most critical to model generalization. Specifically, we propose \textit{Checkpoints Across Time} (\textbf{CAT}): a metric that leverages the variability in perplexity across early checkpoints to identify such critical data points when training MT models. We use this to rank data points, selecting the instances with the highest variability and continuing to train the model solely on this subset. This scoring is used to remove the data points estimated to be least important, thus creating gains in efficiency for training. 

\textbf{CAT} is motivated by fundamental work in machine learning which has shown that there are distinctive periods to learning in deep neural network training. Prior work shows that more common, easier features are learned earlier in training, with the most challenging features learned in the last stages of training \citep{DBLP:journals/corr/abs-2002-03206,DBLP:journals/corr/abs-1711-08856,siddiqui2022metadata}. Prior work has also shown that variance in gradients with respect to inputs~\citep{Agarwal_2022_CVPR} or gradient patterns themselves vary depending on the period of learning~\citep{DBLP:journals/corr/abs-2007-04532}. However, gradients remain expensive to leverage as a learning signal to identify relevant instances. Furthermore, almost the entirety of this work has focused on a computer vision setting, with the exception of~\citet{swayamdipta-etal-2020-dataset} which uses differences in the model's confidence in the true class, and the variability of this confidence across epoch to cluster data. Here – we explore applying the insights that there are distinctive periods of training to an NLP setting and leveraging for the purpose of pruning. We propose a more efficient method to compute signal relative to prior methods which have required computing gradients instead. Our hypothesis is that those with the largest variability indicate the easy examples which show the quickest learning and gains in certainty early in training. 

Compared to existing data pruning strategies in NMT, e.g. the popular (dual) cross-entropy method~\citep{axelrod-etal-2011-domain,junczys-dowmunt-2018-dual} and contrastive data selection~\citep{wang-etal-2018-denoising}, our approach doesn't require an initial clean dataset to train on as this is one limitation that often prevents the adoption of such methods to low-resourced languages. Additionally, it does not require significant time to run as we rely only on two early checkpoints rather than a training run over multiple epochs. This makes \textbf{CAT} methods suitable for low-compute and data-scarce environments which are usually co-occurring~\citep{ahia2021lowresource}. Additionally, it eliminates dependency on other models that may not capture information from various languages and domains.  

We study the effectiveness of \textbf{CAT} across diverse linguistic contexts. Specifically, we utilize English-German and English-French pairs from the Indo-European family using WMT datasets, and the English-Swahili pair from the Bantu family with automatically aligned datasets. Through a series of extensive experiments, we show that \textbf{CAT} results in significant improvements in translation quality over random pruning (randomly selecting $X\%$ subset of the training data) and other embedding-based quality estimation methods such as LaBSE \citep{feng-etal-2022-language}. We summarize our key contributions below:

\begin{enumerate}
    \item Focusing on Machine Translation, we perform an empirical evaluation of existing quality estimation techniques and embedding models as a means of data pruning. These techniques were not designed for data pruning however we observe that quality estimation techniques are less effective for pruning high-quality MT datasets compared to embedding models used for automatic alignment. 
    \item We propose a novel method, \emph{Checkpoints Across Time} (\textbf{CAT}), that leverages perplexity variation to prune MT datasets with significant improvements over existing methods for Indo-European languages. We show that \textbf{CAT} results in significant improvements in translation quality over random pruning (randomly selecting $X\%$ subset of the training data) and other embedding-based quality estimation methods such as LaBSE \citep{feng-etal-2022-language}. Using our best CAT-based method on English-German, English-French and English-Swahili translation tasks, we are able to achieve comparable performance while pruning up to 50\% of the training data. 
    \item We complement our findings with extensive analysis, studying the relationships between performance metrics and sentence characteristics such as sentence length in MT.
\end{enumerate}

\section{Background}\label{sec:background}

\subsection{Data Pruning in ML}
Data pruning has been extensively studied in the work of \cite{marion2023more,10.1007/978-3-030-05677-3_10, Raju2021AcceleratingDL, swayamdipta-etal-2020-dataset,boubdir2023prompts}. \citet{10.1007/978-3-030-05677-3_10}; \citet{sorscher2023neural}  and \citet{Raju2021AcceleratingDL} focus on data pruning for computer vision whereas \citet{swayamdipta-etal-2020-dataset} focuses on data pruning for language tasks. Various techniques have been proposed for data pruning, such as utilization of loss profile \citep{siddiqui2022metadata}, gradients confidence score \citep{Agarwal_2022_CVPR}, and data classification using a model \citep{swayamdipta-etal-2020-dataset}. Increasingly techniques have been extended from primarily computer vision settings \citep{sorscher2023neural} to more recently expanding the treatment of data pruning language settings \citep{marion2023more}. These methods quantify the importance of data points to model generalization. Our work follows along these lines of identifying critical data points by leveraging the difference in perplexity between early checkpoints.

\subsection{Model-based Data Pruning for NMT} 
Specifically for the task of NMT, model-based data pruning approaches have been most popular and originate from the cross-entropy scoring method, proposed by \citet{moore-lewis-2010-intelligent}. Data points are ranked either based on their perplexity under an in-domain language model (LM), or based on the difference of this in-domain cross-entropy and the cross-entropy of a general-domain LM. \citet{axelrod-etal-2011-domain} first apply this technique for MT data selection with a bilingual extension with LM cross-entropy scores for source and target sides. \citet{duh-etal-2013-adaptation} use early neural LMs instead of n-gram LMs, and \citet{junczys-dowmunt-2018-dual} replace LMs with NMT models trained in both directions. \citet{zhang-etal-2020-parallel-corpus} propose to use scores from pre-trained large language models like GPT for domain filtering instead of custom-trained models. \citet{van-der-wees-etal-2017-dynamic} make the data selection dynamic, by gradually fine-tuning an NMT model on newly selected data points every epoch. While the focus of model-based data selection was originally on in-domain data selection, it was later extended to generally filtering noisy data sources. \citet{wang-etal-2018-denoising} propose an online data denoising approach, measuring noise through differences in log probabilities between noisy and denoised NMT models. Most recently, MT quality estimation (QE) models have also been leveraged for corpus mining and filtering \citep{kocyigit-etal-2022-better, batheja-bhattacharyya-2023-little}. \citet{chimoto-bassett-2022-comet} show that COMET-QE \citep{rei-etal-2020-unbabels} improve over random data selection in active learning MT settings.

All these approaches \textit{rely on the existence of a small trusted data set or trusted models} to extrinsically define what ``good data'' looks like. Our approach, while also using model perplexity as a metric, however, does not require such pre-selection, and focuses on model training dynamics instead. Thereby, it can be applied more broadly. Our experiments aim at this less constrained setting without any given high-quality data, so we compare against model-based selection with an MT-QE model and a multilingual LLM. 

\subsection{Embedding-based Data Pruning for NMT}
Embedding-based similarity metrics have recently become popular for NMT data selection, as they rely on unsupervised learning and \textit{intrinsic notions of quality}, assuming that high-quality parallel sentences have high similarity in a cross-lingual representation space~\citep{schwenk-2018-filtering}. Few examples of language-agnostic sentence embedders include MUSE \citep{lample2018unsupervised}, XLM-R \citep{conneau-etal-2020-unsupervised}, LaBSE \citep{feng-etal-2022-language} and LASER \citep{heffernan-etal-2022-bitext}. \citet{schwenk-etal-2021-wikimatrix,schwenk-etal-2021-ccmatrix,nllbteam2022} utilize the LASER sentence embedder for bitext mining, whereas \citet{ramesh-etal-2022-samanantar,indictrans2} use the LaBSE sentence embedder for the same purpose. In our experiments, we compare against data pruning with LASER and LaBSE embeddings.

\subsection{Comparison Studies} 
With this plethora of filtering methods and application scenarios, there have been few works to systematically compare them. The WMT shared tasks for corpus filtering \citep{koehn-etal-2018-findings, koehn-etal-2019-findings, koehn-etal-2020-findings} have shed light on how different languages and resourcefulness conditions pose different challenges for above-described filtering methods in practice, and how they affect downstream NMT performance.
\citet{bane-zaretskaya-2021-selecting} compare data filtering techniques with a more \textit{qualitative} angle: They find that NMT scores and MUSE embeddings have the highest correlation with human quality judgments, NMT scores work best for in-domain selection, and MUSE/XLM-R for out-of-domain generalization. \citet{herold-etal-2022-detecting} compare multiple data noise detection techniques according to their effectiveness for filtering out specific types of data noise. They find that LASER does not detect incorrect languages, and both LASER and cross-entropy perform weakly on detecting misaligned sentences and over/under translation.
Similarly, \citet{bane-etal-2022-comparison} find cross-entropy filtering empirically superior to other methods (XLM-R, MUSE, LASER, COMET) for filtering out various synthetic noise types~\citep{khayrallah-koehn-2018-impact}. In their setup, COMET fails at filtering out misaligned sentences but shows particular sensitivity for target-side omissions, NMT scoring for source-side omissions, while LASER and COMET do not filter mismatching numbers well. \citet{dakwale-etal-2022-empirical} compare LASER and LaBSE on low-resource corpus filtering tasks and find them competitive if languages are included in the pretraining of the underlying embedding models.

In this work, we provide both a systematic empirical comparison of different types of data pruning techniques on downstream NMT performance for three target languages, as well as qualitative analyses of what kind of data is getting selected.

\section{CAT: \emph{Checkpoints Across Time}} 

Our goal is to extract the most valuable data points that would contribute to model generalization and train the model solely on this. Our approach, while simple, leverages the difference in learning stages in deep neural network optimization. Firstly, we train an NMT model on the full dataset for only a few epochs, ideally a small fraction of the total time needed for the model to converge. Subsequently, we use the checkpoints from this initial stage to compute the perplexity of each data point. These perplexity scores serve as the basis for selecting candidate data points for subsequent pruning using different perplexity profiles. This approach draws upon the findings of \citet{swayamdipta-etal-2020-dataset} that categorize training data relevance based on the learning dynamics during training, and \citet{Agarwal_2022_CVPR} which leverage the difference in gradients across training to identify easy versus challenging examples. Unlike \citet{Agarwal_2022_CVPR} which uses computationally expensive variance of gradients, we use perplexity. For CAT we can utilize the perplexity scores in two ways:

\subsection{CAT-DIFF}

Let $X = \{x_1,x_2,x_3,...,x_n\}$ be all instances in the training data. To select $s \subseteq X$ instances to be pruned, we compute the perplexity of each instance at epoch $j$ of all data points in $X$ using the  model trained on $X$ with weight parameters $\theta_j$: 

\begin{equation}
\text{PPL}_j(x_i) = \exp\left(-\frac{1}{T}\sum_{t=1}^{T}\log P_{{\theta}_j}(x_{i,t})\right) 
\end{equation}

where $T$ is the number of words in each example $x_i$. 

We then rank all data points based on differences in perplexity between the two epochs that form our checkpoints:
\begin{equation}
\Delta \text{PPL}(x_i) = \text{PPL}_1(x_i) - \text{PPL}_k(x_i)
\end{equation}
From the ranked data points, we select those with the highest differences between early epochs,\footnote{Either 5th and 1st or 2nd and 1st.} according to a threshold at a chosen percentile. For example, for pruning of 50\% of data points, we keep all data points that rank higher than the 50th percentile.

\subsection{CAT-VAR}

An alternative is to compute the perplexity at $N>1$ epoch checkpoints, but to rank data points according to the variance of the perplexity across the checkpoints: 
\begin{equation}
    \text{Var}(\text{PPL}(x_i)) = \frac{1}{N} \sum_{j=1}^{N} \left(\text{PPL}_j(x_i) - \frac{1}{N} \sum_{l=1}^{N} \text{PPL}_l(x_l)\right)^2
\end{equation}

In our experiments we use $N = 3$ corresponding to epochs $1,3$ and $5$. 
 
In contrast to CAT-DIFF we then select data points around the median with a range according to the pruning percentage that we want to achieve, e.g. data points between the 25th and 75th percentile for a pruning percentage of 50\%. This aligns with several works which have found variance in model signal to be predictive of difficulty \citep{swayamdipta-etal-2020-dataset,Agarwal_2022_CVPR}. \citet{swayamdipta-etal-2020-dataset} show that data points that fall on the tails of the variance tend to be either too easy or too difficult and thus the model may not benefit most from being trained on such examples.

\section{Experimental Setup}
To fully test the CAT methods for pruning we explore the following variations: 
(1) different target languages, (2) different pruning levels, (3) different test domains, and (4) configurations such as model sizes and choice of checkpoints. We compare the effectiveness of the pruning against embedding-based, LLM-based, and random selection.

\subsection{Training Data}
We conduct our experiments on datasets that translate from English into three languages: French, German, and Swahili. For German, we use the English-German WMT19 dataset \citep{barrault-etal-2019-findings}, comprising \smallsim38M parallel sentences sourced from various origins, and for Swahili, the English-Swahili WMT22 dataset \citep{nllbteam2022,adelani-etal-2022-findings}, consisting of \smallsim22M parallel sentences sourced from the web \citep{banon-etal-2020-paracrawl} then automatically identified using LASER. We use the English-French WMT15  \citep{bojar-etal-2015-findings} dataset for French. Due to compute constraints, the majority of our experiments and ablations were carried out on a randomly selected subset of \smallsim3.8M parallel sentences from all datasets, which constitutes 10\% of the original German and French datasets and 17\% of the Swahili dataset. Subsequently, we implement the best variant of our method on the entire dataset to evaluate the scalability of our approach.

\subsection{Evaluation Data}
Our evaluation sets consist of WMT18~\citep{bojar-etal-2018-findings}, WMT15~\cite{bojar-etal-2015-findings}, FLORES \citep{goyal-etal-2022-flores,nllbteam2022} and MAFAND-MT~\citep{adelani-etal-2022-thousand} test sets. These test sets were curated using human annotators thereby mitigating common issues associated with automatically aligned datasets, such as erroneous alignments. The WMT18 and WMT15 test sets were released as part of the WMT machine translation task consisting of 2998 and 1500 parallel sentences respectively from various sources of the news domain. WMT18 was used to evaluate German models while WMT15 test set was used to evaluate French. The FLORES test set was used to evaluate all languages: German, French and Swahili. The FLORES test set consists of 1012 sentences from 3 sources: WikiNews, WikiJunior and WikiVoyage. We use the MAFAND-MT test set for Swahili model evaluation. MAFAND-MT test set consists of 1835 sentences from the news domain.

\subsection{Training Details}

We train all our NMT models using the fairseq library\footnote{\url{https://github.com/facebookresearch/fairseq}} \citep{ott2019fairseq}. Our models follow the same architecture as the vanilla transformer \citep{DBLP:journals/corr/VaswaniSPUJGKP17} with GeLU activation \citep{DBLP:journals/corr/HendrycksG16} instead of ReLU activation \citep{Nair2010RectifiedLU}. We report all the hyperparameters used for training the models in \Cref{tab:hyperparam_details}. We conduct all our experiments on 4 A100 40GB GPUs with the duration for runs at  90\%, 70\%, and 50\% prune level, and full dataset utilization approximately taking 1, 2, 3, and 4.5 hours, respectively. Furthermore, we train separate SentencePiece tokenizers \citep{kudo-richardson-2018-sentencepiece} for source and target languages with a vocab size of 16K using 3.8M translation pairs for both English-German and English-Swahili experiments. We use SacreBLEU library\footnote{With parameters: \texttt{nrefs:1|case:mixed|eff:no|tok:13a|smooth:exp|version:2.3.2}} \citep{post-2018-call} to compute BLEU score \citep{papineni-etal-2002-bleu} and used it as our evaluation metric. We also calculate chrF++ and COMET scores and report them in the Appendix. All the metrics exhibit the same trend in our experiments. Our baseline models are trained on the entire 3.8M translation pairs and denoted as ``\texttt{full}'' for all language pairs; English-German, English-French and English-Swahili. We investigate the impact of data pruning with sparsity levels of 50\%, 70\% and 90\% using various techniques briefly described below.

\subsection{Baselines}

As outlined in Section~\ref{sec:background}, we compare with the most common NMT data pruning techniques:
\begin{enumerate}
    \item \textbf{Random Selection:} This involves choosing data instances without specific criteria purely by chance. Although this approach may seem intuitively suboptimal, both \citet{azeemi-etal-2023-data} and \citet{marion2023more} show that this also performs competitively with different pruning strategies for a few cases. Random selection is also consistently reported in other problems evaluating the ability of a method to identify a salient subset such as weight pruning \citep{Gale2019TheSO} or interpretability \citep{hooker2019benchmark}. Therefore, we include this approach as a meaningful expected lower bound of performance for any pruning method proposed. This helps us answer the question, \textit{is this method better than randomly guessing what data to keep?}
    \item \textbf{COMET-QE:} \citet{rei-etal-2020-unbabels} proposed model-based MT evaluation metric that utilizes embeddings from XLM-RoBERTa \cite{conneau-etal-2020-unsupervised} to access translation quality when provided source, labels and target translations or just source and target translations. 
    \item \textbf{LaBSE \& LASER:} Both are multilingual sentence embedding models trained on large aligned corpora covering a number of languages. LASER \citep{heffernan-etal-2022-bitext} covers 200 languages and was trained with an encoder-decoder LSTM architecture. LaBSE \citep{feng-etal-2022-language} on the other hand covers 93 languages and was trained on top of BERT \cite{devlin-etal-2019-bert}. 
    \item \textbf{BLOOM LLM:} Using the multilingual pre-trained BLOOM LM~\citep{bloom}, we explore the selection of relevant data points by computing the perplexity of the reference sentence.\footnote{Past work has found that NMT is more susceptible to target-side noise~\citep{khayrallah-koehn-2018-impact}, so we focus on target-side perplexity. In principle, this method could be extended to a combination of source and target perplexities as in the cross-entropy method~\citep{moore-lewis-2010-intelligent}, but it would require twice the computation.} This is the closest to the CAT method, as it also leverages insights from a trained Transformer model, where we expect easy and hard-to-learn examples to fall on the tails and thus will not enable the model to learn effectively. In contrast to CAT, its perplexity scores might be more expressive, because it was trained on more data and with more parameters. On the other hand, we do not get any insights on training dynamics. We use BLOOM  variants with 560M, 1.1B and 1.7B parameters for our experiments and assess the impact of model size on perplexity computation.
\end{enumerate}
In the following, we will summarize embedding-based selection and QE-based selection as \emph{Translation Quality Estimators}.

\begin{table}[t]
\small
\centering
\resizebox{0.6\columnwidth}{!}{
\begin{tabular}{lr}
\toprule
Hyperparameter & Value \\
\midrule
Optimizer & Adam \citep{kingma2014adam} \\
Beta values $(\beta_1, \beta_2)$ & $(0.9, 0.98)$ \\
Learning rate & $5 \times 10^{-4}$ \\
Scheduler & Inverse sqrt \\
Criterion & Cross-entropy \\
Label smoothing \citep{labelsmoothing} & $0.1$ \\
Warmup learning rate &  $10^{-7}$ \\
Warmup steps & $4,000$ \\
Gradient clipping & $1.0$ \\
Dropout fraction \citep{dropout} & $0.2$ \\
Effective batch size & $16\mathrm{K}$ \\
Mixed precision training & FP16 \\
Maximum epochs & $30$ \\
Maximum sequence length & $256$ \\
\bottomrule
\end{tabular}
}
\caption{Model hyperparameter settings for all our experiments.}
\label{tab:hyperparam_details}
\end{table}

\section{Results and Discussion}

\begin{figure*}[ht]
  \centering
     \begin{subfigure}[t]{1\textwidth}
         \centering
         \includegraphics[width=8cm]{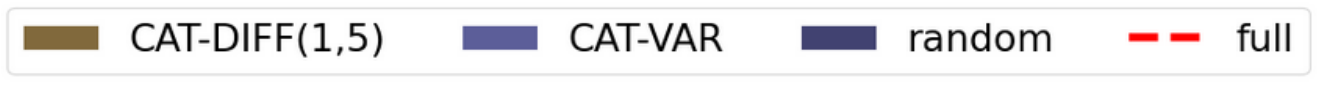}
\end{subfigure}
  \begin{subfigure}[b]{0.45\textwidth}
    \includegraphics[width=\linewidth]{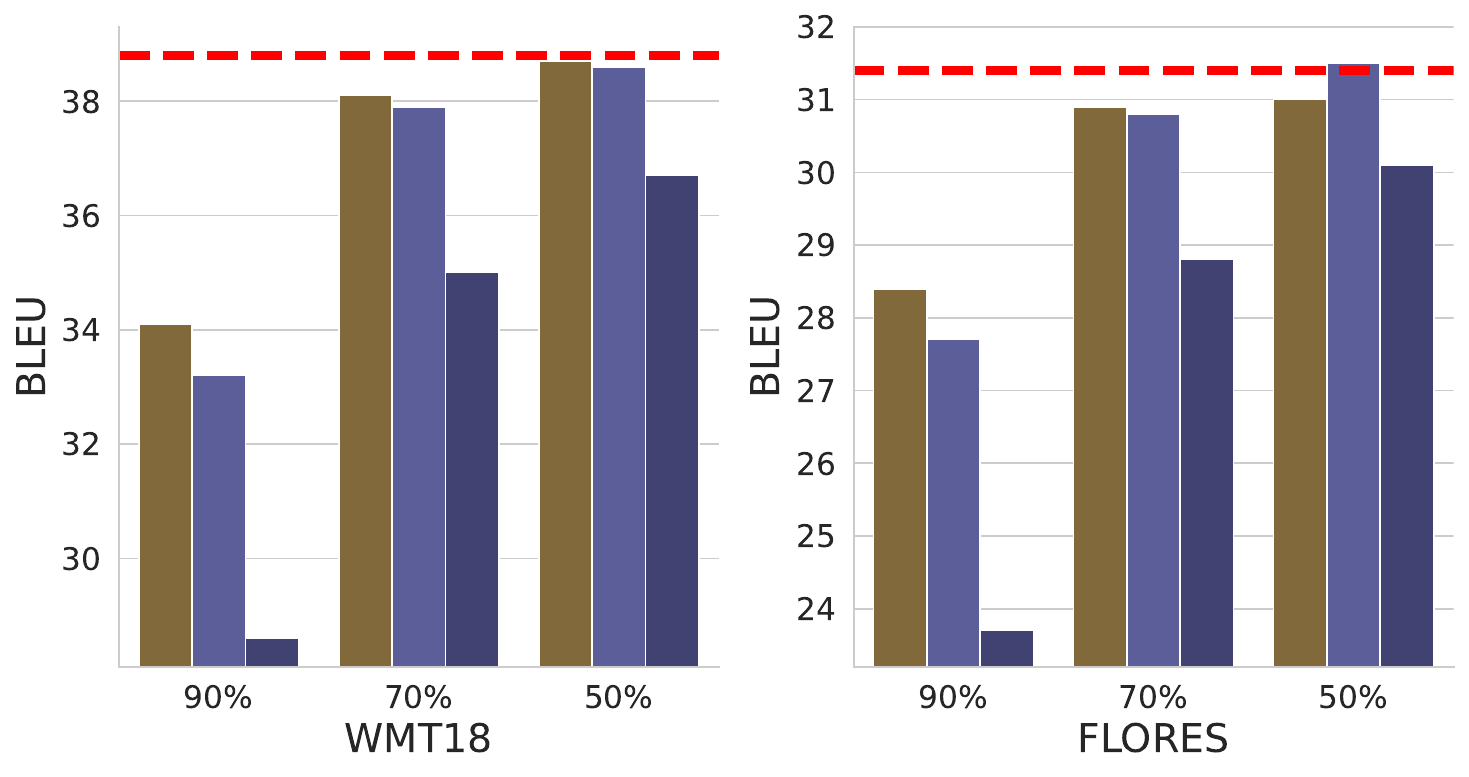}
    \caption{German perplexity}
    \label{fig:PPL_de}
  \end{subfigure}
  \begin{subfigure}[b]{0.45\textwidth}
    \includegraphics[width=\linewidth]{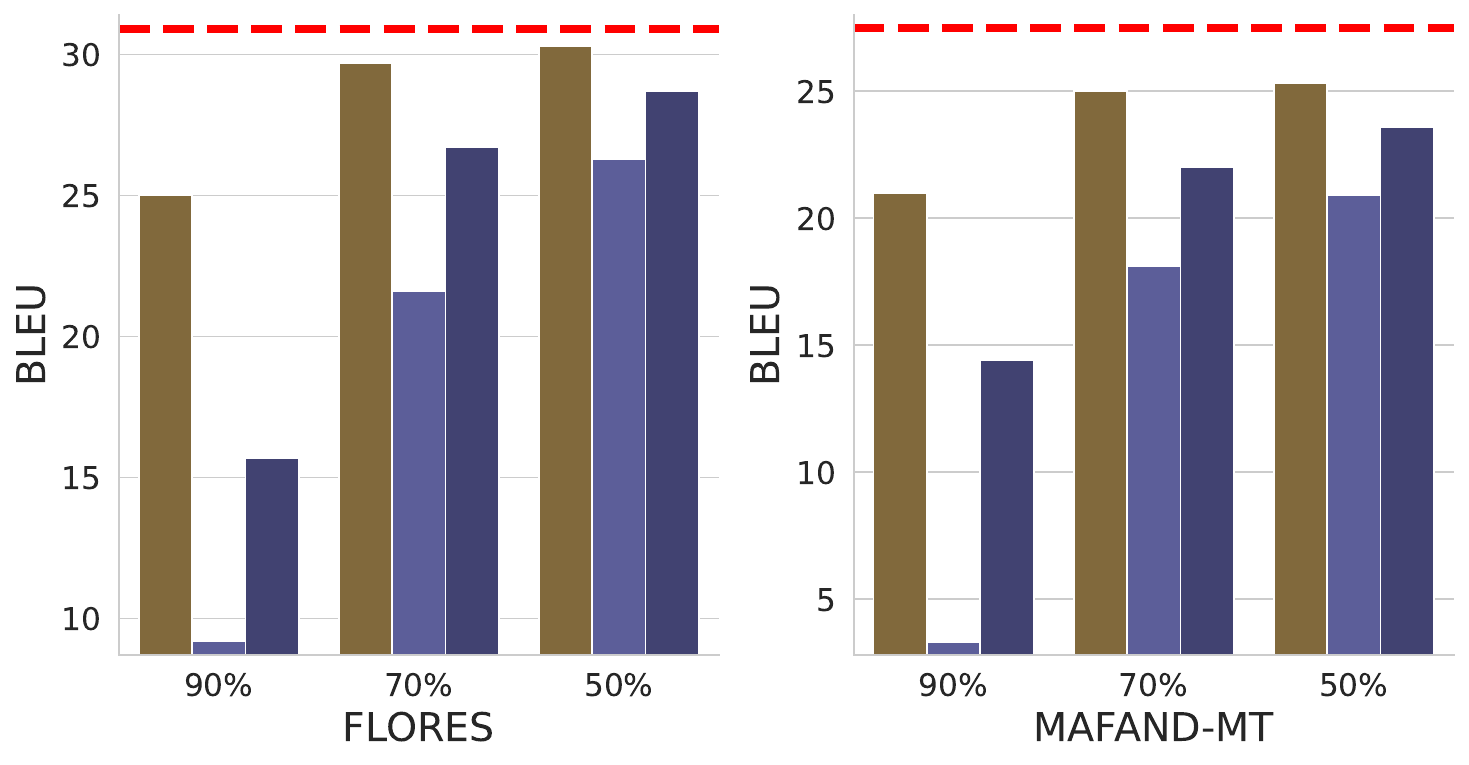}
    \caption{Swahili Perplexity}
    \label{fig:PPL_sw}
  \end{subfigure}
  \caption{Performance of the CAT methods for both German and Swahili. The CAT-DIFF(1,5), referring to the difference between checkpoints 1 and 5, consistently performs better than random. CAT-VAR yields very poor performance for Swahili and also underperforms CAT-DIFF for German.}
  \label{fig:PPL}
\end{figure*}

We conducted experiments on English-German and English-Swahili datasets, employing various pruning techniques. We pruned 90\%, 70\% and 50\% of the training data and compared this to training on the full set. To evaluate these models, we used WMT18 and FLORES test sets for German, and FLORES and MAFAND-MT as the test sets for Swahili. Furthermore, we applied the most effective pruning techniques to an English-French dataset to verify the consistency of performance across different language pairs.

\subsection{CAT techniques outperforms random pruning in German and Swahili}
Figure \ref{fig:PPL_de} demonstrates that CAT-DIFF(1,5) and CAT-VAR outperform random selection in the German test sets. On the other hand, in Swahili, only CAT-DIFF(1,5) consistently outperforms random selection, while selecting data randomly yields better results than CAT-VAR (refer to Figure \ref{fig:PPL_sw}). CAT techniques achieve efficient pruning while maintaining an upwards of 75\% performance compared to training on the full dataset, even after pruning 90\% of the data. For German, both CAT-DIFF(1,5) and CAT-VAR maintain over 80\% of the performance achieved by training on 100\% of the data. In the case of Swahili, CAT-DIFF(1,5) retains above 80\% of the performance, whereas CAT-VAR tends to perform worse, retaining only around 30\% of the performance. A noteworthy efficiency of CAT techniques lies in their ability to sidestep the inferencing with fully converged models, thus one doesn't need a lot of compute to utilize these techniques.

\Cref{tab:catdiffabl} presents the ablation study conducted for CAT-DIFF to compare the performance of CAT pruning techniques based on different checkpoints. We evaluate the difference between checkpoints 1 and 5, referred to as CAT-DIFF(1,5) and checkpoints 1 and 2, denoted as CAT-DIFF(1,2). We see that CAT-DIFF(1,5) yields better performance than CAT-DIFF(1,2). However, CAT-DIFF(1,2) offers the advantage of using fewer resources albeit at the cost of a slight dip in performance.

\subsection{Translation quality estimators are  less effective for German}
\begin{figure*}[ht]
  \centering
  \begin{subfigure}[t]{1\textwidth}
         \centering
         \includegraphics[width=8cm]{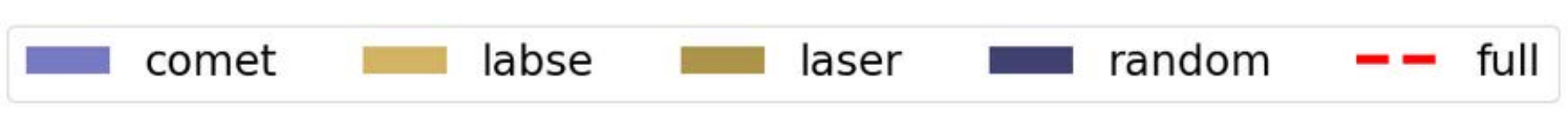}
\end{subfigure}

  \begin{subfigure}[b]{0.45\textwidth}
    \includegraphics[width=\linewidth]{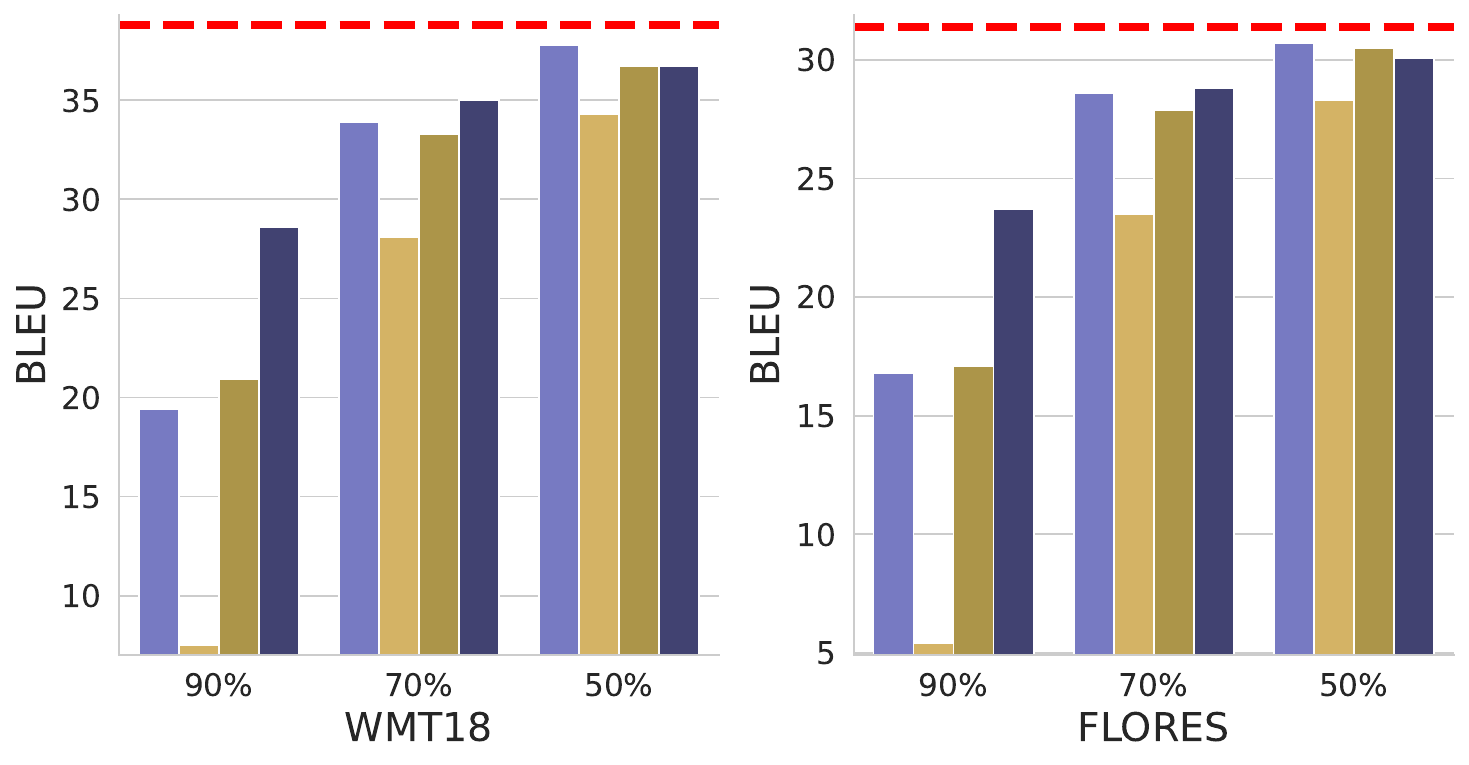}
    \caption{German Translation Quality}
    \label{fig:QET_de}
  \end{subfigure}
  \begin{subfigure}[b]{0.45\textwidth}
    \includegraphics[width=\linewidth]{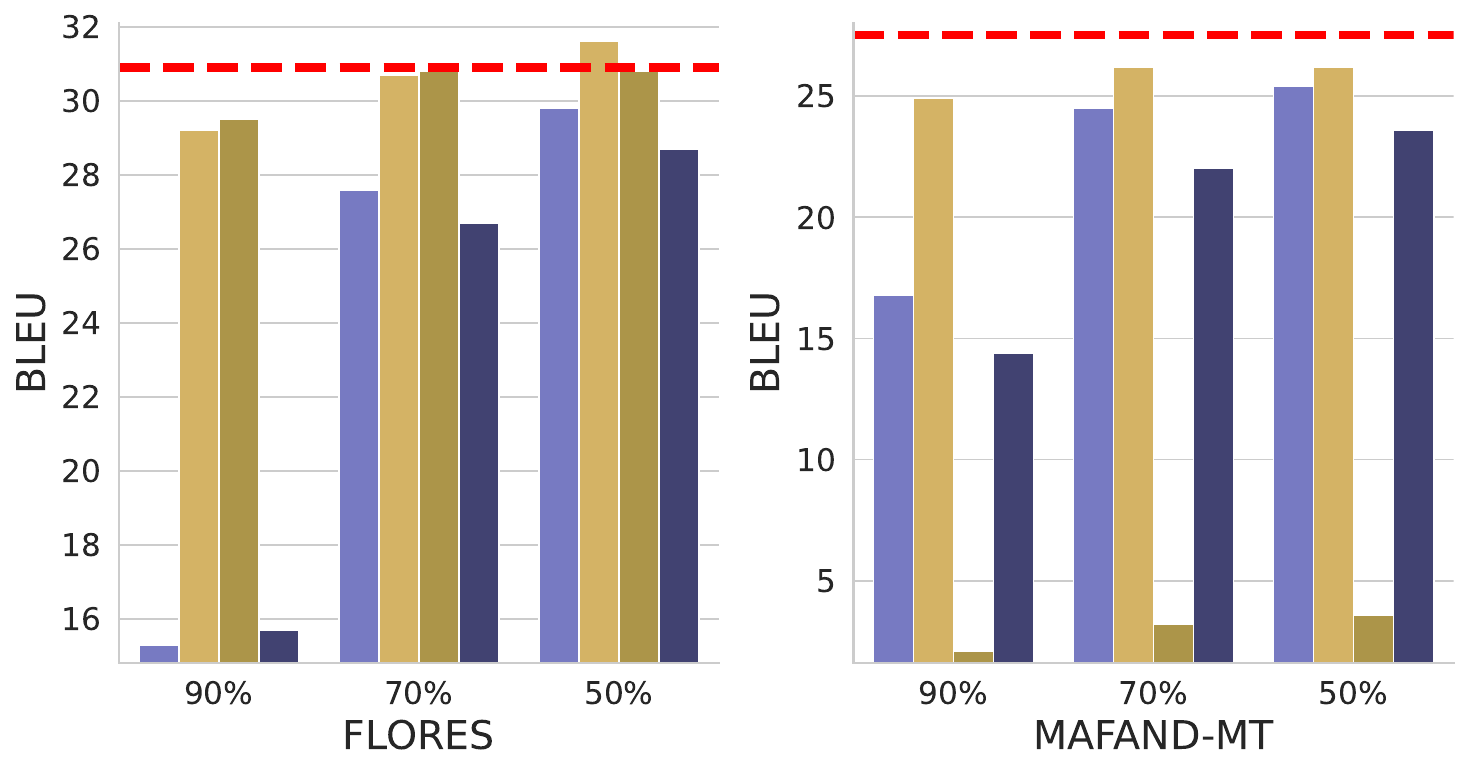}
    \caption{Swahili Translation Quality}
    \label{fig:QET_sw}
  \end{subfigure}
  \caption{Performance of the translation quality metrics for both German and Swahili. Quality estimation metrics perform better than random in an automatically aligned setting (Swahili) than in the high-quality setting (German).}
  \label{fig:QET}
\end{figure*}

We find translation quality estimators to be less effective in pruning German than Swahili. For instance, \Cref{fig:QET_de} indicates that random data pruning outperforms translation quality estimators at 90\% and 70\% pruning levels for both test sets. However, with Swahili, we observe a different behavior, where LABSE outperforms random pruning on all pruning levels,  as depicted in \Cref{fig:QET_sw}. Interestingly, pruning Swahili data with LaBSE and LASER yields superior performance compared to training on the full data set, particularly for the 50\% pruning level for the FLORES test set. Further, translation quality estimators consistently outperform random pruning in Swahili and do not sacrifice performance even at 90\% pruning level with the exception of LASER on the MAFAND test set. We believe that these differences in behavior can be attributed to varying data quality between the two languages. The German dataset contains sentences with high variance in sentence length and included sentences with foreign language. Conversely, the Swahili dataset consists of processed automatically aligned sentences, where there is a low sentence length variance and the sentences do not contain any foreign language. See \Cref{fig:Length} and \Cref{fig:ID} in the Appendix. Therefore, selecting the best sentences ensures the inclusion of high-quality translation pairs.

\begin{table}
\centering
\resizebox{0.7\textwidth}{!}{
\begin{tabular}{lclllllllll}
\toprule
&  & \multicolumn{3}{c}{\textbf{WMT18}} & \multicolumn{3}{c}{\textbf{FLORES}} & \multicolumn{3}{c}{\textbf{MAFAND-MT}} \\ 
\cmidrule{3-5} \cmidrule{6-8} \cmidrule{9-11}
&  & \textit{90\%} & \textit{70\%} & \textit{50\%} & \textit{90\%} & \textit{70\%} & \textit{50\%} & \textit{90\%} & \textit{70\%} & \textit{50\%}\\
\midrule
\multirow{2}{*}{\textbf{German}} & \textsc{CAT-DIFF(1,5)} & \textbf{34.1} & \textbf{38.1} & \textbf{38.7} & \textbf{28.4} & \textbf{30.9} & \textbf{31.0} \\
 & \textsc{CAT-DIFF(1,2)} & 33.4 & 37.7 & 38.1 & 27.7 & 30.1 & 30.9 \\
\midrule
\multirow{2}{*}{\textbf{Swahili}} & \textsc{CAT-DIFF(1,5)} &  &  &  & \textbf{25.0} & 29.7 & \textbf{30.3} & \textbf{21.0} & \textbf{25.0} & 25.3\\
 & \textsc{CAT-DIFF(1,2)} & & & & 23.7 & \textbf{30.1} & 30.0 & 20.6 & 24.6 & \textbf{25.6} \\
\bottomrule
\end{tabular}
}
\caption{Ablation study of the CAT-DIFF techniques for both German and Swahili. CAT-DIFF(1,5) outperforms CAT-DIFF(1,2) except for Swahili at 70\%  and 50\% prune level for FLORES and MAFAND-MT respectively. As a result we use CAT-DIFF(1,5) in all  experiments.}
\label{tab:catdiffabl}
\end{table}

\subsection{LLM perplexity techniques are competitive for Swahili but sub-par for German}
\begin{figure*}[ht]
  \centering
     \begin{subfigure}[t]{1\textwidth}
         \centering
         \includegraphics[width=10cm]{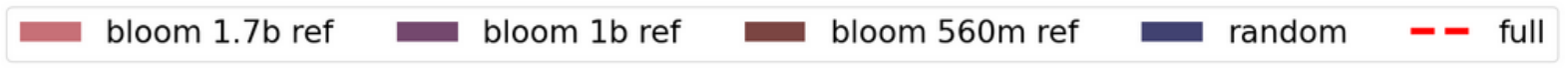}
\end{subfigure}
  \begin{subfigure}[b]{0.45\textwidth}
    \includegraphics[width=\linewidth]{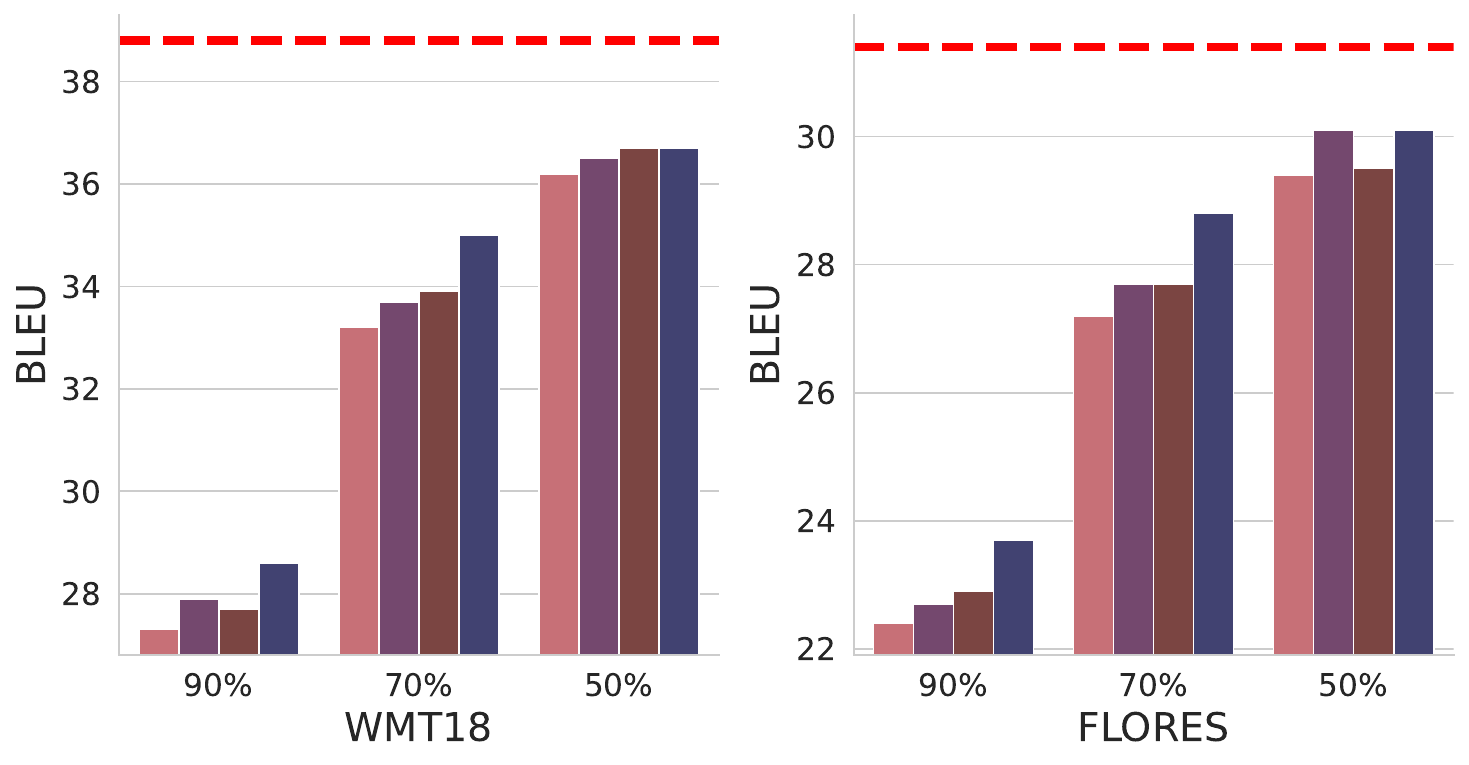}
    \caption{German LLM}
    \label{fig:BLOOM_de}
  \end{subfigure}
  \begin{subfigure}[b]{0.45\textwidth}
    \includegraphics[width=\linewidth]{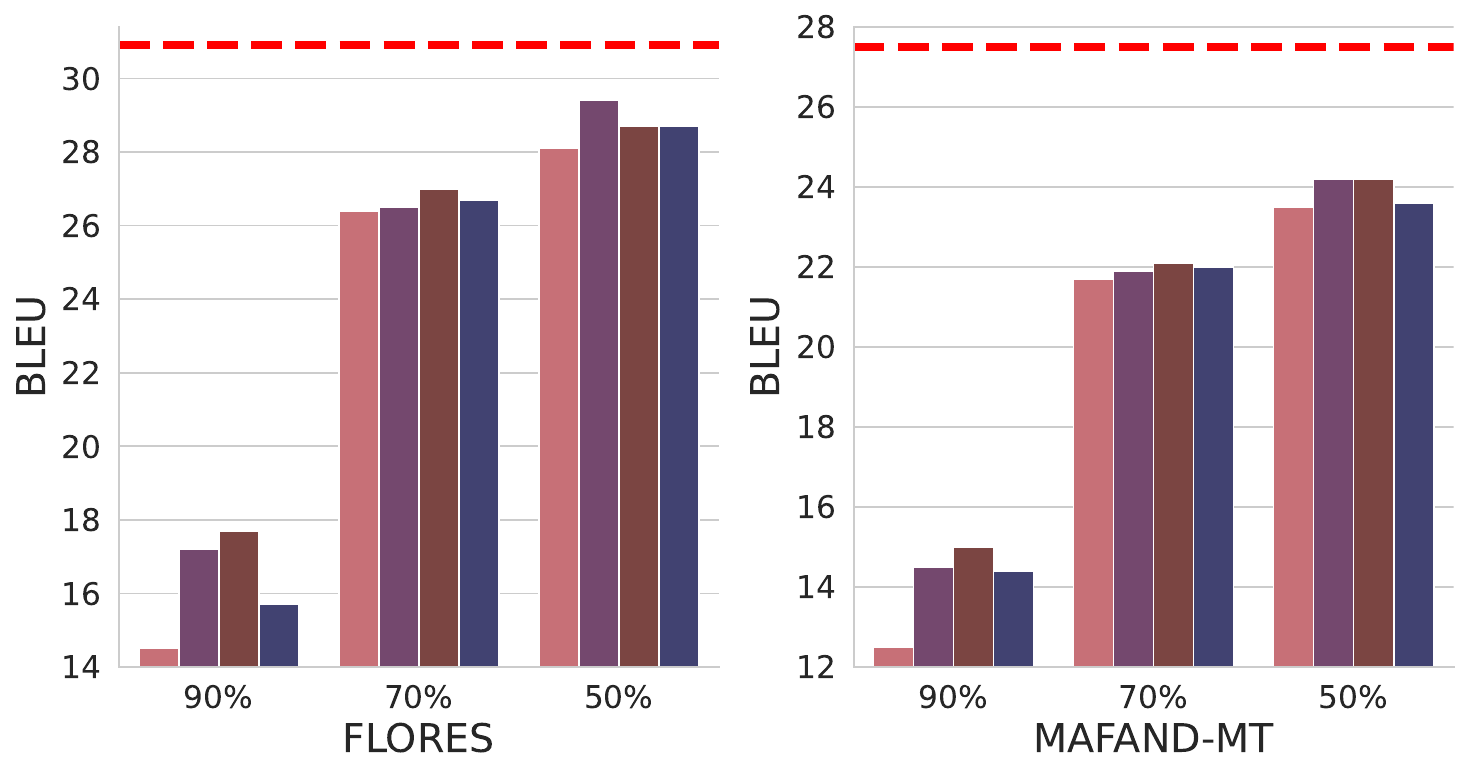}
    \caption{Swahili LLM}
    \label{fig:BLOOM_sw}
  \end{subfigure}
  \caption{Performance of the LLM perplexity metrics for both German and Swahili. For German, the LLM perplexity metrics perform worse than random whereas, for Swahili, the LLM perplexity metrics are approximately the same as random.}
  \label{fig:BLOOM}
\end{figure*}

\Cref{fig:BLOOM} illustrates that utilizing LLMs for pruning provides competitive results for Swahili but shows sub-par performance for German. We further observe that the LLM perplexity technique across different BLOOM model sizes performs superior compared to random pruning across pruning levels for Swahili (see \Cref{fig:BLOOM_sw}). Among the various BLOOM model sizes, the smallest model, BLOOM 560M consistently outperforms the others, followed by the BLOOM 1B model. It is important to note that LLM techniques perform worse than random for German. This suggests that in cases where limitations in data and compute double bind, utilizing pretrained LLMs can save on the limited resources available.

\subsection{What pruning techniques work best?}
\begin{figure}[!h]
  \centering
     \begin{subfigure}[t]{1\textwidth}
         \centering
         \includegraphics[width=8cm]{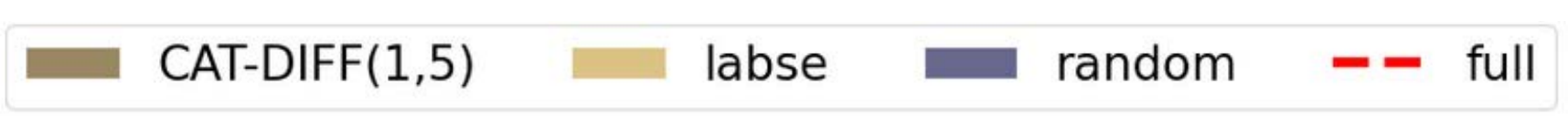}
\end{subfigure}
  \begin{subfigure}[b]{0.3\textwidth}
    \includegraphics[width=\linewidth]{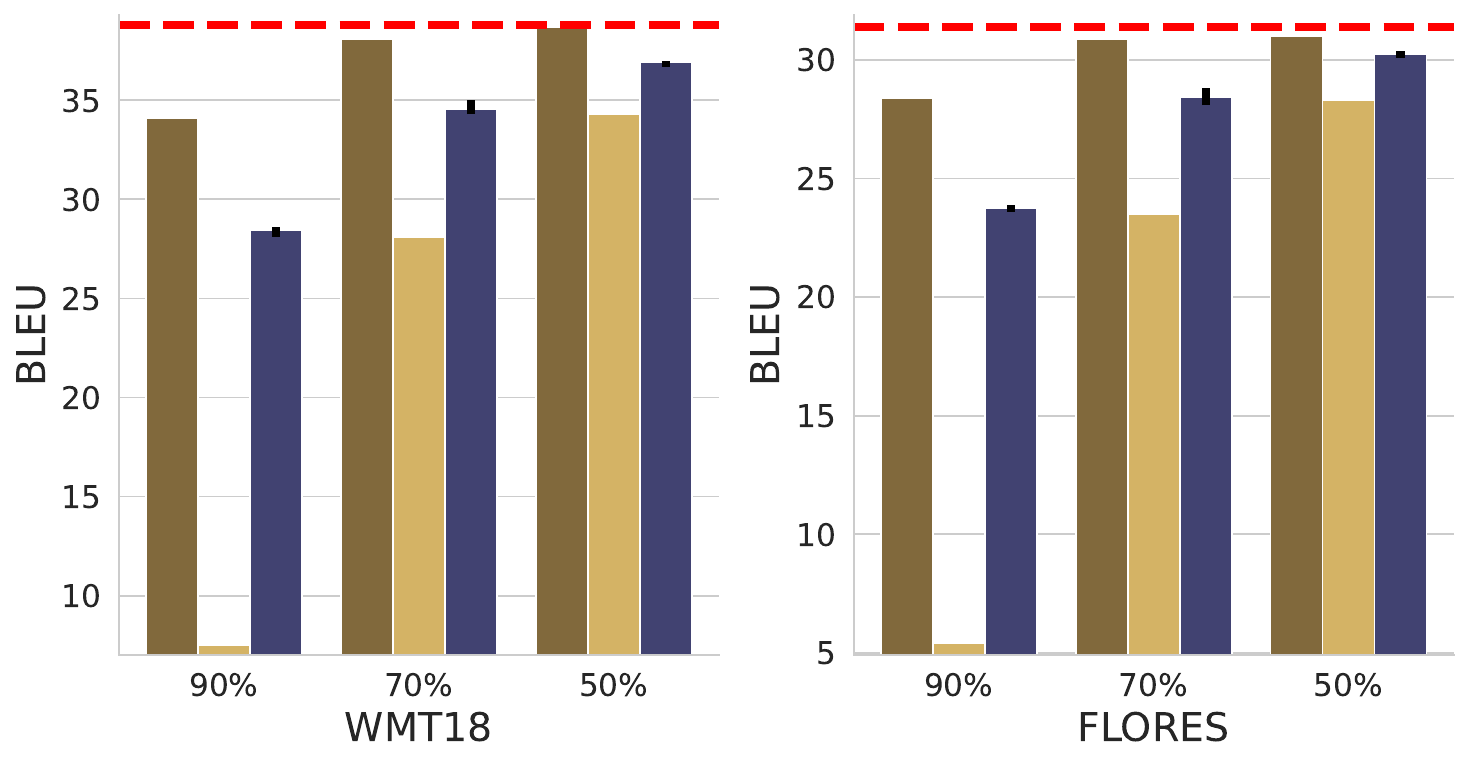}
    \caption{German}
    \label{fig:best_de}
  \end{subfigure}
  \begin{subfigure}[b]{0.3\textwidth}
    \includegraphics[width=\linewidth]{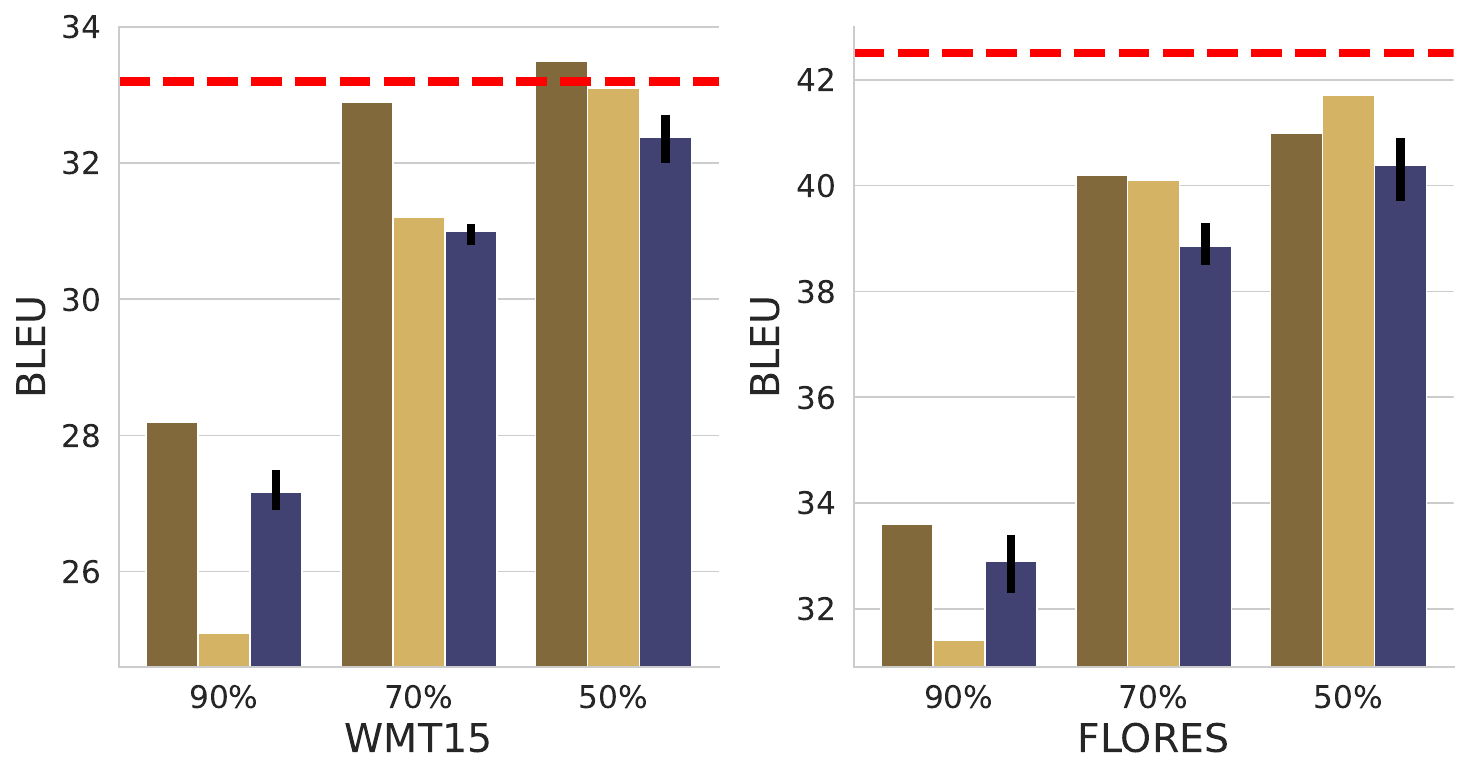}
    \caption{French}
    \label{fig:best_fr}
  \end{subfigure}
  \begin{subfigure}[b]{0.3\textwidth}
    \includegraphics[width=\linewidth]{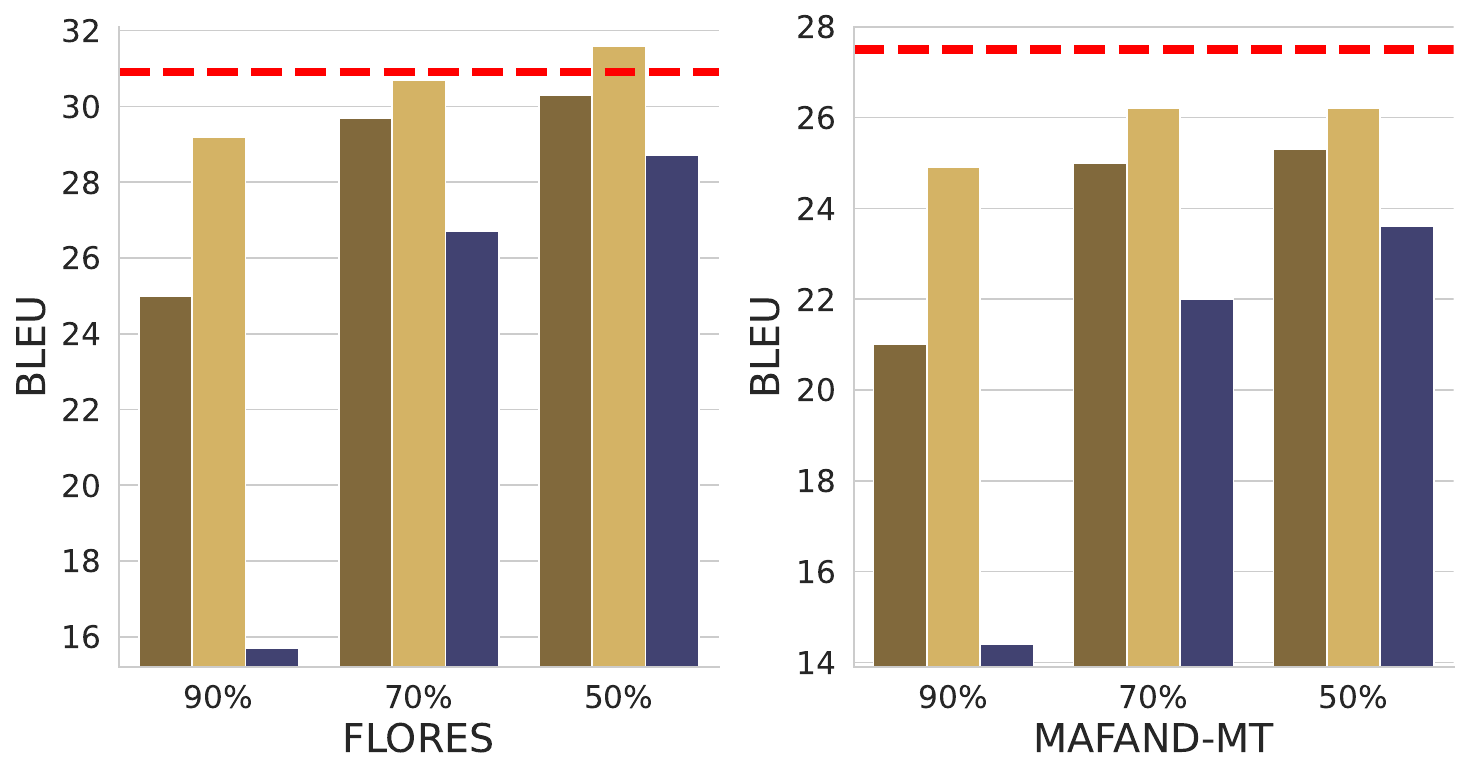}
    \caption{Swahili}
    \label{fig:best_sw}
  \end{subfigure}
  \caption{Best performing techniques against the performance of the full dataset. For both German and French, CAT-DIFF(1,5) achieves above 77\% of the performance of training on the full set for the 90\% prune level. At 50\% pruning, CAT-DIFF(1,5) achieves above 92\% of the full training performance. The error bars on the random samples for German and French plots represent the minimum and maximum BLEU scores from 5 runs with different random seeds for the data selection.}
  \label{fig:best}
\end{figure}

\Cref{fig:best} shows the comparison of random, LaBSE and CAT-DIFF pruning techniques for German, French and Swahili. It is evident that CAT-DIFF, outperforms other methods for German, whereas LaBSE demonstrates superior performance for Swahili. Specifically, CAT-DIFF(1,5) surpasses the other techniques for German at each pruning level and retains 99\% of the full data's performance even at 50\% prune level (refer to \Cref{fig:best_de}). We observe similar trends for French in \Cref{fig:best_fr} with CAT-DIFF(1,5) also generally outperforms LaBSE and random selection, except for FLORES at the 50\% prune level. Conversely, for Swahili, LaBSE is consistently superior to other pruning techniques, followed by CAT-DIFF(1,5) (see \Cref{fig:best_sw}). We also find that pruning using LaBSE for Swahili yields competitive performance as training on the full dataset.

We also report the statistical significance results with paired bootstrap resampling\citep{koehn-2004-statistical} for SacreBLEU and COMET in the Appendix, see \Cref{tab:german,tab:swahili,tab:french}.

\subsection{Scaling best pruning techniques} 
To verify the generalizability of CAT-DIFF along with baselines such as LaBSE and Random Selection, we trained NMT models on the entire dataset, $\sim$23M parallel sentences for Swahili and $\sim$38M parallel sentences for German and French. We compare CAT-DIFF(1,5), LABSE and random data pruning at 90\% prune level. We find the results to be similar to 3.8M scale experiments. \Cref{tab:full_results} show that CAT-DIFF(1,5) outperforms random in German and French while pruning using LaBSE performs better in Swahili. 

\begin{table}[ht]
\small
\centering
\resizebox{0.6\textwidth}{!}{
\begin{tabular}{llcccc}
\toprule
Language & Test Set & Random & LaBSE & \textsc{CAT-DIFF(1,5)} & Full \\
\midrule
\multirow{2}{*}{German} & WMT18 & 38.8 & 9.2 & 40.3 & 42.6 \\
                        & FLORES & 31.4 & 6.4  & 32.1 & 34.3 \\
\midrule
\multirow{2}{*}{Swahili} & FLORES & 29.3 & 33.6 & 20.3 & 33.0 \\
                        & MAFAND-MT & 24.5 & 28.9  & 15.3 & 27.5 \\
\midrule
\multirow{2}{*}{French} & WMT15 & 33.2 & 29.2 & 34.7 & 35.8 \\
                        & FLORES & 42.5 & 36.2  & 43.6 & 45.1 \\
\bottomrule
\end{tabular}
}
\caption{BLEU scores for German, French and Swahili on the entire datasets (i.e. $\sim$38M for German and French and $\sim$23M for Swahili). Random and \textsc{CAT-DIFF} are performed with 10\% sparsity (i.e. 90\% prune level).}
\label{tab:full_results}
\end{table}

\section{Analysis of Sentence Selection for German, French, and Swahili}
\begin{figure}
	\centering
    \resizebox{\columnwidth}{!}{
    \includegraphics[width=10\textwidth]{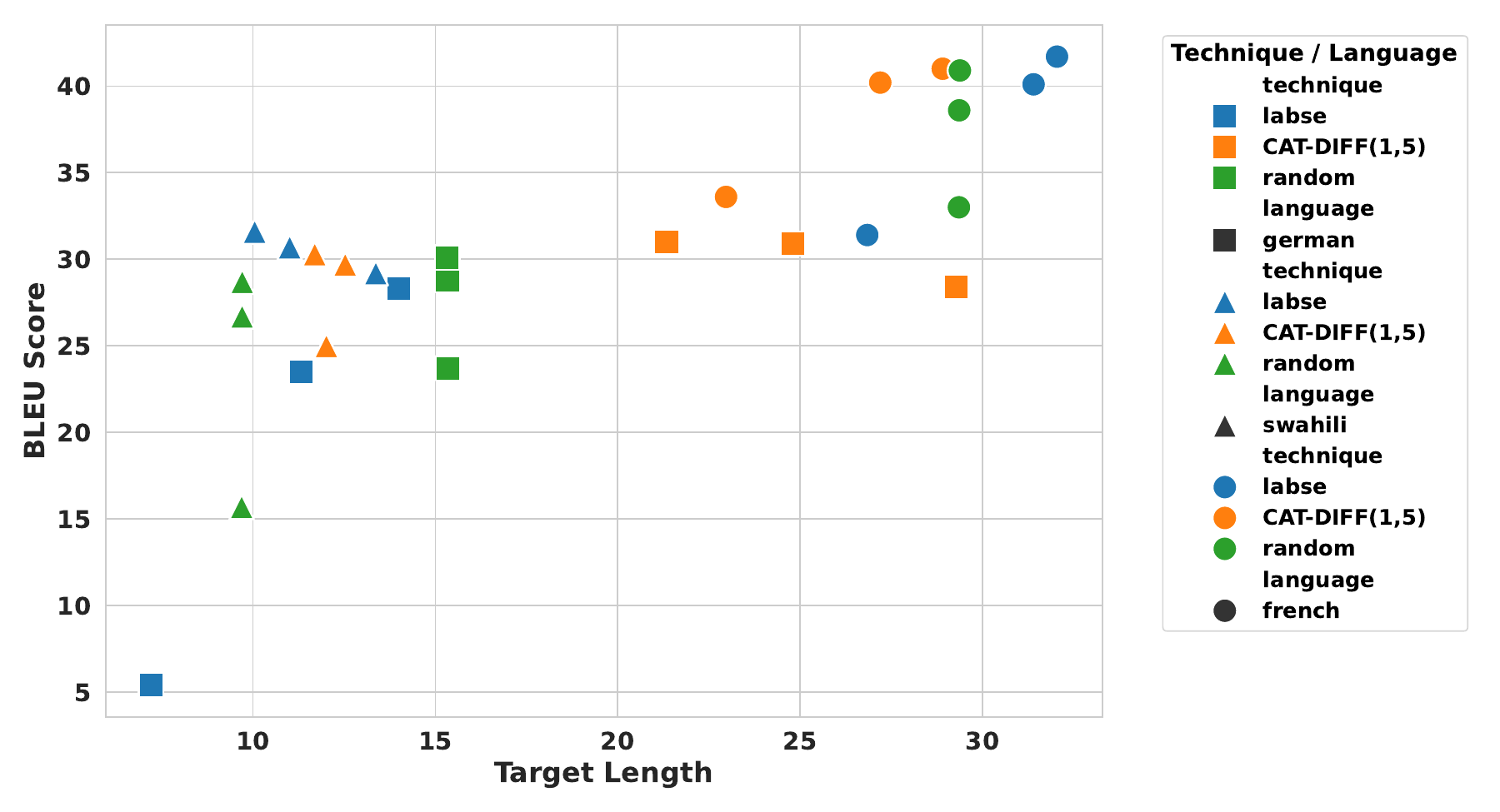}
    }
    \caption{Target sentence length vs BLEU score for all languages and techniques. We see that longer sentences correlate on average with better performance. Notably, pruning techniques like CAT-DIFF(1,5), which favor longer sentences, achieve higher BLEU scores.}
  \label{fig:sent}
\end{figure}

To better understand the performance gap between pruning techniques for German, French(Indo-European languages) and Swahili(a Bantu language), we analyze the sentences selected randomly as well as those selected using CAT-DIFF(1,5) and LaBSE. \Cref{fig:sent} illustrates that longer sentences correlate with higher BLEU scores for German and French when considering target sentence length. This observation of longer sequence lengths giving better scores is also seen in \citet{junczys-dowmunt-2019-microsoft}. Most of the points in the top right quadrant are chosen by CAT-DIFF(1,5), indicating a preference for longer sentences and performs better. However, we observe that picking the longest sentences for Swahili doesn't lead to better BLEU scores. We, therefore also look at the lexical diversity to distinguish the sentences selected.

\begin{figure}
	\centering
    \resizebox{\columnwidth}{!}{
    \includegraphics[width=10\textwidth]{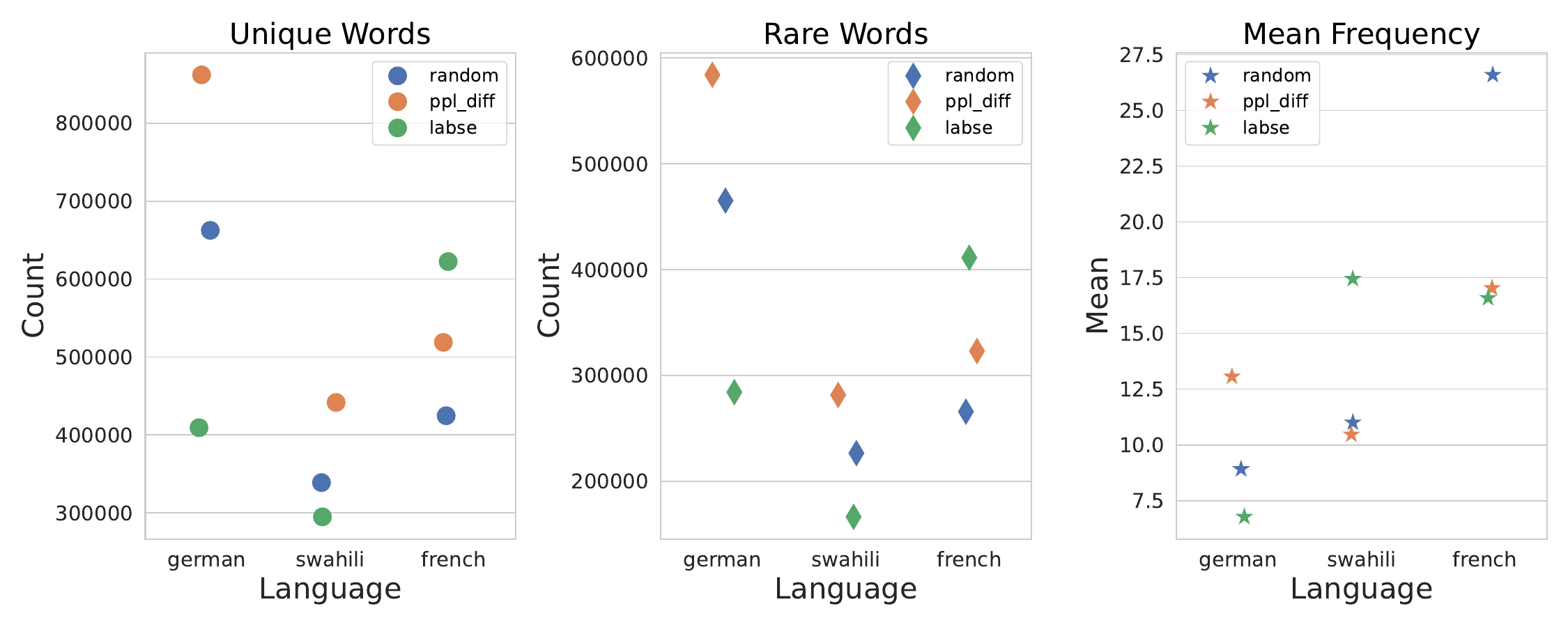}
    }
    \caption{Analysis of word characteristics selected by different pruning techniques. CAT-DIFF(1,5) tends to select more unique and rare words for German and Swahili and the words selected are characterized by a lower mean word frequency. In contrast, for French, LaBSE favors unique and rare words, and its selection is less pronounced compared to CAT-DIFF(1,5)'s for German.  These statistics illustrate that pruning techniques prioritize nuanced examples across languages without adhering to a uniform pattern.}
  \label{fig:word_analysis}
\end{figure}

\Cref{fig:word_analysis} shows that CAT-DIFF(1,5) selects sentences that have more unique and rare words as opposed to German which is less lexically diverse and thus results in lower mean frequency. Similarly, CAT-DIFF(1,5) also selects more unique and rare words for Swahili. Although LaBSE performs relatively better in selecting more unique and rare words for French, it still falls short compared selection of words using CAT-DIFF(1,5) for German. This implies that CAT-DIFF(1,5) selection is far more complex than selecting the longest sentences. We leave a deeper investigation of sentences selected using various techniques for future studies.

\section{Conclusions}

Identifying optimal subsets of a given dataset for training a machine translation model is a crucial problem that enables data efficiency and saves on compute. In this paper, we present \textit{Checkpoints Across Time} (CAT) methods as efficient techniques to prune datasets. CAT uses the change in perplexity scores of individual examples over the first few training epochs to find good subsets. Using 3.8M sentences for English-German, English-French and English-Swahili, we show that CAT techniques offer highly computationally efficient pruning that achieves above 75\% of full performance even when 90\% pruning is applied to all target languages. We also show that LaBSE, when used to select training examples from an automatically aligned dataset provides strong signals on examples crucial for model generalization as the examples selected perform on par or surpass training on the full dataset. Interestingly, we find that utilizing LLMs for pruning does not offer benefits in selecting data for German but does for Swahili.

\section{Limitations}

In this work, we focus on data pruning in a bilingual setting for 3 languages and limit the data size to 3.8M sentences. This is due to computation constraints. This means that the results may not apply to other languages. Also, it would be interesting to investigate performance on larger scales as we only conducted experiments on German and English for CAT-DIFF and random selection.  Moreover, further research is needed to explore the multilingual setup where data pruning would be conducted simultaneously on several languages.

\section{Acknowledgements}

EAC acknowledges a grant from the Carnegie Corporation of New York (provided through the African Institute for Mathematical Sciences). The statements made and views expressed are solely the responsibility of the authors.

\newpage

\bibliography{anthology,egbib}

\begin{thebibliography}{82}
\providecommand{\natexlab}[1]{#1}
\providecommand{\url}[1]{\texttt{#1}}
\expandafter\ifx\csname urlstyle\endcsname\relax
  \providecommand{\doi}[1]{doi: #1}\else
  \providecommand{\doi}{doi: \begingroup \urlstyle{rm}\Url}\fi

\bibitem[Abdulmumin et~al.(2022)Abdulmumin, Beukman, Alabi, Emezue, Chimoto, Adewumi, Muhammad, Adeyemi, Yousuf, Singh, and Gwadabe]{abdulmumin-etal-2022-separating}
Idris Abdulmumin, Michael Beukman, Jesujoba Alabi, Chris~Chinenye Emezue, Everlyn Chimoto, Tosin Adewumi, Shamsuddeen Muhammad, Mofetoluwa Adeyemi, Oreen Yousuf, Sahib Singh, and Tajuddeen Gwadabe.
\newblock Separating grains from the chaff: Using data filtering to improve multilingual translation for low-resourced {A}frican languages.
\newblock In \emph{Proceedings of the Seventh Conference on Machine Translation (WMT)}, pp.\  1001--1014, Abu Dhabi, United Arab Emirates (Hybrid), December 2022. Association for Computational Linguistics.
\newblock URL \url{https://aclanthology.org/2022.wmt-1.98}.

\bibitem[Achille et~al.(2017)Achille, Rovere, and Soatto]{DBLP:journals/corr/abs-1711-08856}
Alessandro Achille, Matteo Rovere, and Stefano Soatto.
\newblock Critical learning periods in deep neural networks.
\newblock \emph{CoRR}, abs/1711.08856, 2017.
\newblock URL \url{http://arxiv.org/abs/1711.08856}.

\bibitem[Adelani et~al.(2022{\natexlab{a}})Adelani, Alabi, Fan, Kreutzer, Shen, Reid, Ruiter, Klakow, Nabende, Chang, Gwadabe, Sackey, Dossou, Emezue, Leong, Beukman, Muhammad, Jarso, Yousuf, Niyongabo~Rubungo, Hacheme, Wairagala, Nasir, Ajibade, Ajayi, Gitau, Abbott, Ahmed, Ochieng, Aremu, Ogayo, Mukiibi, Ouoba~Kabore, Kalipe, Mbaye, Tapo, Memdjokam~Koagne, Munkoh-Buabeng, Wagner, Abdulmumin, Awokoya, Buzaaba, Sibanda, Bukula, and Manthalu]{adelani-etal-2022-thousand}
David Adelani, Jesujoba Alabi, Angela Fan, Julia Kreutzer, Xiaoyu Shen, Machel Reid, Dana Ruiter, Dietrich Klakow, Peter Nabende, Ernie Chang, Tajuddeen Gwadabe, Freshia Sackey, Bonaventure F.~P. Dossou, Chris Emezue, Colin Leong, Michael Beukman, Shamsuddeen Muhammad, Guyo Jarso, Oreen Yousuf, Andre Niyongabo~Rubungo, Gilles Hacheme, Eric~Peter Wairagala, Muhammad~Umair Nasir, Benjamin Ajibade, Tunde Ajayi, Yvonne Gitau, Jade Abbott, Mohamed Ahmed, Millicent Ochieng, Anuoluwapo Aremu, Perez Ogayo, Jonathan Mukiibi, Fatoumata Ouoba~Kabore, Godson Kalipe, Derguene Mbaye, Allahsera~Auguste Tapo, Victoire Memdjokam~Koagne, Edwin Munkoh-Buabeng, Valencia Wagner, Idris Abdulmumin, Ayodele Awokoya, Happy Buzaaba, Blessing Sibanda, Andiswa Bukula, and Sam Manthalu.
\newblock A few thousand translations go a long way! leveraging pre-trained models for {A}frican news translation.
\newblock In \emph{Proceedings of the 2022 Conference of the North American Chapter of the Association for Computational Linguistics: Human Language Technologies}, pp.\  3053--3070, Seattle, United States, July 2022{\natexlab{a}}. Association for Computational Linguistics.
\newblock \doi{10.18653/v1/2022.naacl-main.223}.
\newblock URL \url{https://aclanthology.org/2022.naacl-main.223}.

\bibitem[Adelani et~al.(2022{\natexlab{b}})Adelani, Alam, Anastasopoulos, Bhagia, Costa-juss{\`a}, Dodge, Faisal, Federmann, Fedorova, Guzm{\'a}n, Koshelev, Maillard, Marivate, Mbuya, Mourachko, Saleem, Schwenk, and Wenzek]{adelani-etal-2022-findings}
David Adelani, Md~Mahfuz~Ibn Alam, Antonios Anastasopoulos, Akshita Bhagia, Marta~R. Costa-juss{\`a}, Jesse Dodge, Fahim Faisal, Christian Federmann, Natalia Fedorova, Francisco Guzm{\'a}n, Sergey Koshelev, Jean Maillard, Vukosi Marivate, Jonathan Mbuya, Alexandre Mourachko, Safiyyah Saleem, Holger Schwenk, and Guillaume Wenzek.
\newblock Findings of the {WMT}{'}22 shared task on large-scale machine translation evaluation for {A}frican languages.
\newblock In \emph{Proceedings of the Seventh Conference on Machine Translation (WMT)}, pp.\  773--800, Abu Dhabi, United Arab Emirates (Hybrid), December 2022{\natexlab{b}}. Association for Computational Linguistics.
\newblock URL \url{https://aclanthology.org/2022.wmt-1.72}.

\bibitem[Agarwal et~al.(2022)Agarwal, D'souza, and Hooker]{Agarwal_2022_CVPR}
Chirag Agarwal, Daniel D'souza, and Sara Hooker.
\newblock Estimating example difficulty using variance of gradients.
\newblock In \emph{Proceedings of the IEEE/CVF Conference on Computer Vision and Pattern Recognition (CVPR)}, pp.\  10368--10378, June 2022.

\bibitem[Ahia et~al.(2021)Ahia, Kreutzer, and Hooker]{ahia2021lowresource}
Orevaoghene Ahia, Julia Kreutzer, and Sara Hooker.
\newblock The low-resource double bind: An empirical study of pruning for low-resource machine translation, 2021.

\bibitem[Arivazhagan et~al.(2019)Arivazhagan, Bapna, Firat, Lepikhin, Johnson, Krikun, Chen, Cao, Foster, Cherry, Macherey, Chen, and Wu]{Arivazhagan2019MassivelyMN}
N.~Arivazhagan, Ankur Bapna, Orhan Firat, Dmitry Lepikhin, Melvin Johnson, Maxim Krikun, Mia~Xu Chen, Yuan Cao, George~F. Foster, Colin Cherry, Wolfgang Macherey, Z.~Chen, and Yonghui Wu.
\newblock Massively multilingual neural machine translation in the wild: Findings and challenges.
\newblock \emph{ArXiv}, abs/1907.05019, 2019.

\bibitem[Arora et~al.(2021)Arora, Tomar, and Agrawal]{arora2021Studying}
Karunesh~Kumar Arora, Geetam~S Tomar, and Shyam~S Agrawal.
\newblock Studying the role of data quality on statistical and neural machine translation.
\newblock In \emph{2021 10th IEEE International Conference on Communication Systems and Network Technologies (CSNT)}, pp.\  199--204. IEEE, 2021.
\newblock URL \url{https://ieeexplore.ieee.org/abstract/document/9509604}.

\bibitem[Axelrod et~al.(2011)Axelrod, He, and Gao]{axelrod-etal-2011-domain}
Amittai Axelrod, Xiaodong He, and Jianfeng Gao.
\newblock Domain adaptation via pseudo in-domain data selection.
\newblock In \emph{Proceedings of the 2011 Conference on Empirical Methods in Natural Language Processing}, pp.\  355--362, Edinburgh, Scotland, UK., July 2011. Association for Computational Linguistics.
\newblock URL \url{https://aclanthology.org/D11-1033}.

\bibitem[Azeemi et~al.(2023)Azeemi, Qazi, and Raza]{azeemi-etal-2023-data}
Abdul Azeemi, Ihsan Qazi, and Agha Raza.
\newblock Data pruning for efficient model pruning in neural machine translation.
\newblock In Houda Bouamor, Juan Pino, and Kalika Bali (eds.), \emph{Findings of the Association for Computational Linguistics: EMNLP 2023}, pp.\  236--246, Singapore, December 2023. Association for Computational Linguistics.
\newblock \doi{10.18653/v1/2023.findings-emnlp.18}.
\newblock URL \url{https://aclanthology.org/2023.findings-emnlp.18}.

\bibitem[Bane \& Zaretskaya(2021)Bane and Zaretskaya]{bane-zaretskaya-2021-selecting}
Fred Bane and Anna Zaretskaya.
\newblock Selecting the best data filtering method for {NMT} training.
\newblock In \emph{Proceedings of Machine Translation Summit XVIII: Users and Providers Track}, pp.\  89--97, Virtual, August 2021. Association for Machine Translation in the Americas.
\newblock URL \url{https://aclanthology.org/2021.mtsummit-up.9}.

\bibitem[Bane et~al.(2022)Bane, Uguet, Stribi{\.z}ew, and Zaretskaya]{bane-etal-2022-comparison}
Fred Bane, Celia~Soler Uguet, Wiktor Stribi{\.z}ew, and Anna Zaretskaya.
\newblock A comparison of data filtering methods for neural machine translation.
\newblock In \emph{Proceedings of the 15th Biennial Conference of the Association for Machine Translation in the Americas (Volume 2: Users and Providers Track and Government Track)}, pp.\  313--325, Orlando, USA, September 2022. Association for Machine Translation in the Americas.
\newblock URL \url{https://aclanthology.org/2022.amta-upg.22}.

\bibitem[Ba{\~n}{\'o}n et~al.(2020)Ba{\~n}{\'o}n, Chen, Haddow, Heafield, Hoang, Espl{\`a}-Gomis, Forcada, Kamran, Kirefu, Koehn, Ortiz~Rojas, Pla~Sempere, Ram{\'\i}rez-S{\'a}nchez, Sarr{\'\i}as, Strelec, Thompson, Waites, Wiggins, and Zaragoza]{banon-etal-2020-paracrawl}
Marta Ba{\~n}{\'o}n, Pinzhen Chen, Barry Haddow, Kenneth Heafield, Hieu Hoang, Miquel Espl{\`a}-Gomis, Mikel~L. Forcada, Amir Kamran, Faheem Kirefu, Philipp Koehn, Sergio Ortiz~Rojas, Leopoldo Pla~Sempere, Gema Ram{\'\i}rez-S{\'a}nchez, Elsa Sarr{\'\i}as, Marek Strelec, Brian Thompson, William Waites, Dion Wiggins, and Jaume Zaragoza.
\newblock {P}ara{C}rawl: Web-scale acquisition of parallel corpora.
\newblock In \emph{Proceedings of the 58th Annual Meeting of the Association for Computational Linguistics}, pp.\  4555--4567, Online, July 2020. Association for Computational Linguistics.
\newblock \doi{10.18653/v1/2020.acl-main.417}.
\newblock URL \url{https://aclanthology.org/2020.acl-main.417}.

\bibitem[Bapna et~al.(2022)Bapna, Caswell, Kreutzer, Firat, van Esch, Siddhant, Niu, Baljekar, Garc{\'i}a, Macherey, Breiner, Axelrod, Riesa, Cao, Chen, Macherey, Krikun, Wang, Gutkin, Shah, Huang, Chen, Wu, and Hughes]{Bapna2022BuildingMT}
Ankur Bapna, Isaac Caswell, Julia Kreutzer, Orhan Firat, Daan van Esch, Aditya Siddhant, Mengmeng Niu, Pallavi~N. Baljekar, Xavier Garc{\'i}a, Wolfgang Macherey, Theresa Breiner, Vera Axelrod, Jason Riesa, Yuan Cao, Mia~Xu Chen, Klaus Macherey, Maxim Krikun, Pidong Wang, Alexander Gutkin, Apurva Shah, Yanping Huang, Z.~Chen, Yonghui Wu, and Macduff Hughes.
\newblock Building machine translation systems for the next thousand languages.
\newblock \emph{ArXiv}, abs/2205.03983, 2022.

\bibitem[Barrault et~al.(2019)Barrault, Bojar, Costa-juss{\`a}, Federmann, Fishel, Graham, Haddow, Huck, Koehn, Malmasi, Monz, M{\"u}ller, Pal, Post, and Zampieri]{barrault-etal-2019-findings}
Lo{\"\i}c Barrault, Ond{\v{r}}ej Bojar, Marta~R. Costa-juss{\`a}, Christian Federmann, Mark Fishel, Yvette Graham, Barry Haddow, Matthias Huck, Philipp Koehn, Shervin Malmasi, Christof Monz, Mathias M{\"u}ller, Santanu Pal, Matt Post, and Marcos Zampieri.
\newblock Findings of the 2019 conference on machine translation ({WMT}19).
\newblock In \emph{Proceedings of the Fourth Conference on Machine Translation (Volume 2: Shared Task Papers, Day 1)}, pp.\  1--61, Florence, Italy, August 2019. Association for Computational Linguistics.
\newblock \doi{10.18653/v1/W19-5301}.
\newblock URL \url{https://aclanthology.org/W19-5301}.

\bibitem[Batheja \& Bhattacharyya(2023)Batheja and Bhattacharyya]{batheja-bhattacharyya-2023-little}
Akshay Batheja and Pushpak Bhattacharyya.
\newblock {``}a little is enough{''}: Few-shot quality estimation based corpus filtering improves machine translation.
\newblock In Anna Rogers, Jordan Boyd-Graber, and Naoaki Okazaki (eds.), \emph{Findings of the Association for Computational Linguistics: ACL 2023}, pp.\  14175--14185, Toronto, Canada, July 2023. Association for Computational Linguistics.
\newblock \doi{10.18653/v1/2023.findings-acl.892}.
\newblock URL \url{https://aclanthology.org/2023.findings-acl.892}.

\bibitem[Bojar et~al.(2015)Bojar, Chatterjee, Federmann, Haddow, Huck, Hokamp, Koehn, Logacheva, Monz, Negri, Post, Scarton, Specia, and Turchi]{bojar-etal-2015-findings}
Ond{\v{r}}ej Bojar, Rajen Chatterjee, Christian Federmann, Barry Haddow, Matthias Huck, Chris Hokamp, Philipp Koehn, Varvara Logacheva, Christof Monz, Matteo Negri, Matt Post, Carolina Scarton, Lucia Specia, and Marco Turchi.
\newblock Findings of the 2015 workshop on statistical machine translation.
\newblock In \emph{Proceedings of the Tenth Workshop on Statistical Machine Translation}, pp.\  1--46, Lisbon, Portugal, September 2015. Association for Computational Linguistics.
\newblock \doi{10.18653/v1/W15-3001}.
\newblock URL \url{https://aclanthology.org/W15-3001}.

\bibitem[Bojar et~al.(2018)Bojar, Federmann, Fishel, Graham, Haddow, Huck, Koehn, and Monz]{bojar-etal-2018-findings}
Ond{\v{r}}ej Bojar, Christian Federmann, Mark Fishel, Yvette Graham, Barry Haddow, Matthias Huck, Philipp Koehn, and Christof Monz.
\newblock Findings of the 2018 conference on machine translation ({WMT}18).
\newblock In \emph{Proceedings of the Third Conference on Machine Translation: Shared Task Papers}, pp.\  272--303, Belgium, Brussels, October 2018. Association for Computational Linguistics.
\newblock \doi{10.18653/v1/W18-6401}.
\newblock URL \url{https://aclanthology.org/W18-6401}.

\bibitem[Boubdir et~al.(2023)Boubdir, Kim, Ermis, Fadaee, and Hooker]{boubdir2023prompts}
Meriem Boubdir, Edward Kim, Beyza Ermis, Marzieh Fadaee, and Sara Hooker.
\newblock Which prompts make the difference? data prioritization for efficient human llm evaluation, 2023.

\bibitem[Chimoto \& Bassett(2022)Chimoto and Bassett]{chimoto-bassett-2022-comet}
Everlyn Chimoto and Bruce Bassett.
\newblock {COMET}-{QE} and active learning for low-resource machine translation.
\newblock In \emph{Findings of the Association for Computational Linguistics: EMNLP 2022}, pp.\  4735--4740, Abu Dhabi, United Arab Emirates, December 2022. Association for Computational Linguistics.
\newblock URL \url{https://aclanthology.org/2022.findings-emnlp.348}.

\bibitem[Conneau et~al.(2020)Conneau, Khandelwal, Goyal, Chaudhary, Wenzek, Guzm{\'a}n, Grave, Ott, Zettlemoyer, and Stoyanov]{conneau-etal-2020-unsupervised}
Alexis Conneau, Kartikay Khandelwal, Naman Goyal, Vishrav Chaudhary, Guillaume Wenzek, Francisco Guzm{\'a}n, Edouard Grave, Myle Ott, Luke Zettlemoyer, and Veselin Stoyanov.
\newblock Unsupervised cross-lingual representation learning at scale.
\newblock In \emph{Proceedings of the 58th Annual Meeting of the Association for Computational Linguistics}, pp.\  8440--8451, Online, July 2020. Association for Computational Linguistics.
\newblock \doi{10.18653/v1/2020.acl-main.747}.
\newblock URL \url{https://aclanthology.org/2020.acl-main.747}.

\bibitem[Costa-juss{\`a} et~al.(2022)Costa-juss{\`a}, Cross, cCelebi, Elbayad, Heafield, Heffernan, Kalbassi, Lam, Licht, Maillard, Sun, Wang, Wenzek, Youngblood, Akula, Barrault, Gonzalez, Hansanti, Hoffman, Jarrett, Sadagopan, Rowe, Spruit, Tran, Andrews, Ayan, Bhosale, Edunov, Fan, Gao, Goswami, Guzm'an, Koehn, Mourachko, Ropers, Saleem, Schwenk, and Wang]{nllbteam2022}
Marta~Ruiz Costa-juss{\`a}, James Cross, Onur cCelebi, Maha Elbayad, Kenneth Heafield, Kevin Heffernan, Elahe Kalbassi, Janice Lam, Daniel Licht, Jean Maillard, Anna Sun, Skyler Wang, Guillaume Wenzek, Alison Youngblood, Bapi Akula, Lo{\"i}c Barrault, Gabriel~Mejia Gonzalez, Prangthip Hansanti, John Hoffman, Semarley Jarrett, Kaushik~Ram Sadagopan, Dirk Rowe, Shannon~L. Spruit, C.~Tran, Pierre~Yves Andrews, Necip~Fazil Ayan, Shruti Bhosale, Sergey Edunov, Angela Fan, Cynthia Gao, Vedanuj Goswami, Francisco Guzm'an, Philipp Koehn, Alexandre Mourachko, Christophe Ropers, Safiyyah Saleem, Holger Schwenk, and Jeff Wang.
\newblock No language left behind: Scaling human-centered machine translation.
\newblock \emph{ArXiv}, abs/2207.04672, 2022.

\bibitem[Dakwale et~al.(2022)Dakwale, Khalil, and Denis]{dakwale-etal-2022-empirical}
Praveen Dakwale, Talaat Khalil, and Brandon Denis.
\newblock Empirical evaluation of language agnostic filtering of parallel data for low resource languages.
\newblock In \emph{Proceedings of the 36th Pacific Asia Conference on Language, Information and Computation}, pp.\  346--355, Manila, Philippines, October 2022. De La Salle University.
\newblock URL \url{https://aclanthology.org/2022.paclic-1.38}.

\bibitem[Devlin et~al.(2019)Devlin, Chang, Lee, and Toutanova]{devlin-etal-2019-bert}
Jacob Devlin, Ming-Wei Chang, Kenton Lee, and Kristina Toutanova.
\newblock {BERT}: Pre-training of deep bidirectional transformers for language understanding.
\newblock In \emph{Proceedings of the 2019 Conference of the North {A}merican Chapter of the Association for Computational Linguistics: Human Language Technologies, Volume 1 (Long and Short Papers)}, pp.\  4171--4186, Minneapolis, Minnesota, June 2019. Association for Computational Linguistics.
\newblock \doi{10.18653/v1/N19-1423}.
\newblock URL \url{https://aclanthology.org/N19-1423}.

\bibitem[Duh et~al.(2013)Duh, Neubig, Sudoh, and Tsukada]{duh-etal-2013-adaptation}
Kevin Duh, Graham Neubig, Katsuhito Sudoh, and Hajime Tsukada.
\newblock Adaptation data selection using neural language models: Experiments in machine translation.
\newblock In \emph{Proceedings of the 51st Annual Meeting of the Association for Computational Linguistics (Volume 2: Short Papers)}, pp.\  678--683, Sofia, Bulgaria, August 2013. Association for Computational Linguistics.
\newblock URL \url{https://aclanthology.org/P13-2119}.

\bibitem[Faghri et~al.(2020)Faghri, Duvenaud, Fleet, and Ba]{DBLP:journals/corr/abs-2007-04532}
Fartash Faghri, David Duvenaud, David~J. Fleet, and Jimmy Ba.
\newblock A study of gradient variance in deep learning.
\newblock \emph{CoRR}, abs/2007.04532, 2020.
\newblock URL \url{https://arxiv.org/abs/2007.04532}.

\bibitem[Fan et~al.(2021)Fan, Bhosale, Schwenk, Ma, El-Kishky, Goyal, Baines, Celebi, Wenzek, Chaudhary, Goyal, Birch, Liptchinsky, Edunov, Auli, and Joulin]{m2m100}
Angela Fan, Shruti Bhosale, Holger Schwenk, Zhiyi Ma, Ahmed El-Kishky, Siddharth Goyal, Mandeep Baines, Onur Celebi, Guillaume Wenzek, Vishrav Chaudhary, Naman Goyal, Tom Birch, Vitaliy Liptchinsky, Sergey Edunov, Michael Auli, and Armand Joulin.
\newblock Beyond english-centric multilingual machine translation.
\newblock \emph{Journal of Machine Learning Research}, 22\penalty0 (107):\penalty0 1--48, 2021.
\newblock URL \url{http://jmlr.org/papers/v22/20-1307.html}.

\bibitem[Feng et~al.(2022)Feng, Yang, Cer, Arivazhagan, and Wang]{feng-etal-2022-language}
Fangxiaoyu Feng, Yinfei Yang, Daniel Cer, Naveen Arivazhagan, and Wei Wang.
\newblock Language-agnostic {BERT} sentence embedding.
\newblock In \emph{Proceedings of the 60th Annual Meeting of the Association for Computational Linguistics (Volume 1: Long Papers)}, pp.\  878--891, Dublin, Ireland, May 2022. Association for Computational Linguistics.
\newblock \doi{10.18653/v1/2022.acl-long.62}.
\newblock URL \url{https://aclanthology.org/2022.acl-long.62}.

\bibitem[Freitag et~al.(2021)Freitag, Foster, Grangier, Ratnakar, Tan, and Macherey]{freitag-etal-2021-experts}
Markus Freitag, George Foster, David Grangier, Viresh Ratnakar, Qijun Tan, and Wolfgang Macherey.
\newblock Experts, errors, and context: A large-scale study of human evaluation for machine translation.
\newblock \emph{Transactions of the Association for Computational Linguistics}, 9:\penalty0 1460--1474, 2021.
\newblock \doi{10.1162/tacl_a_00437}.
\newblock URL \url{https://aclanthology.org/2021.tacl-1.87}.

\bibitem[Gala et~al.(2023)Gala, Chitale, Raghavan, Gumma, Doddapaneni, M, Nawale, Sujatha, Puduppully, Raghavan, Kumar, Khapra, Dabre, and Kunchukuttan]{indictrans2}
Jay Gala, Pranjal~A Chitale, A~K Raghavan, Varun Gumma, Sumanth Doddapaneni, Aswanth~Kumar M, Janki~Atul Nawale, Anupama Sujatha, Ratish Puduppully, Vivek Raghavan, Pratyush Kumar, Mitesh~M Khapra, Raj Dabre, and Anoop Kunchukuttan.
\newblock Indictrans2: Towards high-quality and accessible machine translation models for all 22 scheduled indian languages.
\newblock \emph{Transactions on Machine Learning Research}, 2023.
\newblock ISSN 2835-8856.
\newblock URL \url{https://openreview.net/forum?id=vfT4YuzAYA}.

\bibitem[Gale et~al.(2019)Gale, Elsen, and Hooker]{Gale2019TheSO}
Trevor Gale, Erich Elsen, and Sara Hooker.
\newblock The state of sparsity in deep neural networks.
\newblock \emph{ArXiv}, abs/1902.09574, 2019.
\newblock URL \url{https://api.semanticscholar.org/CorpusID:67855585}.

\bibitem[Goyal et~al.(2022)Goyal, Gao, Chaudhary, Chen, Wenzek, Ju, Krishnan, Ranzato, Guzm{\'a}n, and Fan]{goyal-etal-2022-flores}
Naman Goyal, Cynthia Gao, Vishrav Chaudhary, Peng-Jen Chen, Guillaume Wenzek, Da~Ju, Sanjana Krishnan, Marc{'}Aurelio Ranzato, Francisco Guzm{\'a}n, and Angela Fan.
\newblock The {F}lores-101 evaluation benchmark for low-resource and multilingual machine translation.
\newblock \emph{Transactions of the Association for Computational Linguistics}, 10:\penalty0 522--538, 2022.
\newblock \doi{10.1162/tacl_a_00474}.
\newblock URL \url{https://aclanthology.org/2022.tacl-1.30}.

\bibitem[Guo et~al.(2018)Guo, Shen, Yang, Ge, Cer, Hernandez~Abrego, Stevens, Constant, Sung, Strope, and Kurzweil]{guo-etal-2018-effective}
Mandy Guo, Qinlan Shen, Yinfei Yang, Heming Ge, Daniel Cer, Gustavo Hernandez~Abrego, Keith Stevens, Noah Constant, Yun-Hsuan Sung, Brian Strope, and Ray Kurzweil.
\newblock Effective parallel corpus mining using bilingual sentence embeddings.
\newblock In \emph{Proceedings of the Third Conference on Machine Translation: Research Papers}, pp.\  165--176, Brussels, Belgium, October 2018. Association for Computational Linguistics.
\newblock \doi{10.18653/v1/W18-6317}.
\newblock URL \url{https://aclanthology.org/W18-6317}.

\bibitem[Heffernan et~al.(2022)Heffernan, {\c{C}}elebi, and Schwenk]{heffernan-etal-2022-bitext}
Kevin Heffernan, Onur {\c{C}}elebi, and Holger Schwenk.
\newblock Bitext mining using distilled sentence representations for low-resource languages.
\newblock In \emph{Findings of the Association for Computational Linguistics: EMNLP 2022}, pp.\  2101--2112, Abu Dhabi, United Arab Emirates, December 2022. Association for Computational Linguistics.
\newblock URL \url{https://aclanthology.org/2022.findings-emnlp.154}.

\bibitem[Hendrycks \& Gimpel(2016)Hendrycks and Gimpel]{DBLP:journals/corr/HendrycksG16}
Dan Hendrycks and Kevin Gimpel.
\newblock Bridging nonlinearities and stochastic regularizers with gaussian error linear units.
\newblock \emph{CoRR}, abs/1606.08415, 2016.
\newblock URL \url{http://arxiv.org/abs/1606.08415}.

\bibitem[Herold et~al.(2022)Herold, Rosendahl, Vanvinckenroye, and Ney]{herold-etal-2022-detecting}
Christian Herold, Jan Rosendahl, Joris Vanvinckenroye, and Hermann Ney.
\newblock Detecting various types of noise for neural machine translation.
\newblock In \emph{Findings of the Association for Computational Linguistics: ACL 2022}, pp.\  2542--2551, Dublin, Ireland, May 2022. Association for Computational Linguistics.
\newblock \doi{10.18653/v1/2022.findings-acl.200}.
\newblock URL \url{https://aclanthology.org/2022.findings-acl.200}.

\bibitem[Hooker et~al.(2019)Hooker, Erhan, Kindermans, and Kim]{hooker2019benchmark}
Sara Hooker, Dumitru Erhan, Pieter-Jan Kindermans, and Been Kim.
\newblock A benchmark for interpretability methods in deep neural networks, 2019.

\bibitem[Jiang et~al.(2020)Jiang, Zhang, Talwar, and Mozer]{DBLP:journals/corr/abs-2002-03206}
Ziheng Jiang, Chiyuan Zhang, Kunal Talwar, and Michael~C. Mozer.
\newblock Exploring the memorization-generalization continuum in deep learning.
\newblock \emph{CoRR}, abs/2002.03206, 2020.
\newblock URL \url{https://arxiv.org/abs/2002.03206}.

\bibitem[Johnson et~al.(2017)Johnson, Schuster, Le, Krikun, Wu, Chen, Thorat, Vi{\'e}gas, Wattenberg, Corrado, Hughes, and Dean]{johnson-etal-2017-googles}
Melvin Johnson, Mike Schuster, Quoc~V. Le, Maxim Krikun, Yonghui Wu, Zhifeng Chen, Nikhil Thorat, Fernanda Vi{\'e}gas, Martin Wattenberg, Greg Corrado, Macduff Hughes, and Jeffrey Dean.
\newblock {G}oogle{'}s multilingual neural machine translation system: Enabling zero-shot translation.
\newblock \emph{Transactions of the Association for Computational Linguistics}, 5:\penalty0 339--351, 2017.
\newblock \doi{10.1162/tacl_a_00065}.
\newblock URL \url{https://aclanthology.org/Q17-1024}.

\bibitem[Junczys-Dowmunt(2018)]{junczys-dowmunt-2018-dual}
Marcin Junczys-Dowmunt.
\newblock Dual conditional cross-entropy filtering of noisy parallel corpora.
\newblock In \emph{Proceedings of the Third Conference on Machine Translation: Shared Task Papers}, pp.\  888--895, Belgium, Brussels, October 2018. Association for Computational Linguistics.
\newblock \doi{10.18653/v1/W18-6478}.
\newblock URL \url{https://aclanthology.org/W18-6478}.

\bibitem[Junczys-Dowmunt(2019)]{junczys-dowmunt-2019-microsoft}
Marcin Junczys-Dowmunt.
\newblock {M}icrosoft translator at {WMT} 2019: Towards large-scale document-level neural machine translation.
\newblock In \emph{Proceedings of the Fourth Conference on Machine Translation (Volume 2: Shared Task Papers, Day 1)}, pp.\  225--233, Florence, Italy, August 2019. Association for Computational Linguistics.
\newblock \doi{10.18653/v1/W19-5321}.
\newblock URL \url{https://aclanthology.org/W19-5321}.

\bibitem[Khayrallah \& Koehn(2018)Khayrallah and Koehn]{khayrallah-koehn-2018-impact}
Huda Khayrallah and Philipp Koehn.
\newblock On the impact of various types of noise on neural machine translation.
\newblock In \emph{Proceedings of the 2nd Workshop on Neural Machine Translation and Generation}, pp.\  74--83, Melbourne, Australia, July 2018. Association for Computational Linguistics.
\newblock \doi{10.18653/v1/W18-2709}.
\newblock URL \url{https://aclanthology.org/W18-2709}.

\bibitem[Kingma \& Ba(2014)Kingma and Ba]{kingma2014adam}
Diederik~P. Kingma and Jimmy Ba.
\newblock Adam: A method for stochastic optimization.
\newblock \emph{International Conference On Learning Representations}, 2014.

\bibitem[Kocyigit et~al.(2022)Kocyigit, Lee, and Wijaya]{kocyigit-etal-2022-better}
Muhammed Kocyigit, Jiho Lee, and Derry Wijaya.
\newblock Better quality estimation for low resource corpus mining.
\newblock In \emph{Findings of the Association for Computational Linguistics: ACL 2022}, pp.\  533--543, Dublin, Ireland, May 2022. Association for Computational Linguistics.
\newblock \doi{10.18653/v1/2022.findings-acl.45}.
\newblock URL \url{https://aclanthology.org/2022.findings-acl.45}.

\bibitem[Koehn(2004)]{koehn-2004-statistical}
Philipp Koehn.
\newblock Statistical significance tests for machine translation evaluation.
\newblock In \emph{Proceedings of the 2004 Conference on Empirical Methods in Natural Language Processing}, pp.\  388--395, Barcelona, Spain, July 2004. Association for Computational Linguistics.
\newblock URL \url{https://aclanthology.org/W04-3250}.

\bibitem[Koehn et~al.(2018)Koehn, Khayrallah, Heafield, and Forcada]{koehn-etal-2018-findings}
Philipp Koehn, Huda Khayrallah, Kenneth Heafield, and Mikel~L. Forcada.
\newblock Findings of the {WMT} 2018 shared task on parallel corpus filtering.
\newblock In \emph{Proceedings of the Third Conference on Machine Translation: Shared Task Papers}, pp.\  726--739, Belgium, Brussels, October 2018. Association for Computational Linguistics.
\newblock \doi{10.18653/v1/W18-6453}.
\newblock URL \url{https://aclanthology.org/W18-6453}.

\bibitem[Koehn et~al.(2019)Koehn, Guzm{\'a}n, Chaudhary, and Pino]{koehn-etal-2019-findings}
Philipp Koehn, Francisco Guzm{\'a}n, Vishrav Chaudhary, and Juan Pino.
\newblock Findings of the {WMT} 2019 shared task on parallel corpus filtering for low-resource conditions.
\newblock In \emph{Proceedings of the Fourth Conference on Machine Translation (Volume 3: Shared Task Papers, Day 2)}, pp.\  54--72, Florence, Italy, August 2019. Association for Computational Linguistics.
\newblock \doi{10.18653/v1/W19-5404}.
\newblock URL \url{https://aclanthology.org/W19-5404}.

\bibitem[Koehn et~al.(2020)Koehn, Chaudhary, El-Kishky, Goyal, Chen, and Guzm{\'a}n]{koehn-etal-2020-findings}
Philipp Koehn, Vishrav Chaudhary, Ahmed El-Kishky, Naman Goyal, Peng-Jen Chen, and Francisco Guzm{\'a}n.
\newblock Findings of the {WMT} 2020 shared task on parallel corpus filtering and alignment.
\newblock In \emph{Proceedings of the Fifth Conference on Machine Translation}, pp.\  726--742, Online, November 2020. Association for Computational Linguistics.
\newblock URL \url{https://aclanthology.org/2020.wmt-1.78}.

\bibitem[Koneru et~al.(2022)Koneru, Liu, and Niehues]{DBLP:journals/corr/abs-2201-05700}
Sai Koneru, Danni Liu, and Jan Niehues.
\newblock Cost-effective training in low-resource neural machine translation.
\newblock \emph{CoRR}, abs/2201.05700, 2022.
\newblock URL \url{https://arxiv.org/abs/2201.05700}.

\bibitem[Kreutzer et~al.(2022)Kreutzer, Caswell, Wang, Wahab, van Esch, Ulzii-Orshikh, Tapo, Subramani, Sokolov, Sikasote, Setyawan, Sarin, Samb, Sagot, Rivera, Rios, Papadimitriou, Osei, Suarez, Orife, Ogueji, Rubungo, Nguyen, M{\"u}ller, M{\"u}ller, Muhammad, Muhammad, Mnyakeni, Mirzakhalov, Matangira, Leong, Lawson, Kudugunta, Jernite, Jenny, Firat, Dossou, Dlamini, de~Silva, {\c{C}}abuk~Ball{\i}, Biderman, Battisti, Baruwa, Bapna, Baljekar, Azime, Awokoya, Ataman, Ahia, Ahia, Agrawal, and Adeyemi]{kreutzer-etal-2022-quality}
Julia Kreutzer, Isaac Caswell, Lisa Wang, Ahsan Wahab, Daan van Esch, Nasanbayar Ulzii-Orshikh, Allahsera Tapo, Nishant Subramani, Artem Sokolov, Claytone Sikasote, Monang Setyawan, Supheakmungkol Sarin, Sokhar Samb, Beno{\^\i}t Sagot, Clara Rivera, Annette Rios, Isabel Papadimitriou, Salomey Osei, Pedro~Ortiz Suarez, Iroro Orife, Kelechi Ogueji, Andre~Niyongabo Rubungo, Toan~Q. Nguyen, Mathias M{\"u}ller, Andr{\'e} M{\"u}ller, Shamsuddeen~Hassan Muhammad, Nanda Muhammad, Ayanda Mnyakeni, Jamshidbek Mirzakhalov, Tapiwanashe Matangira, Colin Leong, Nze Lawson, Sneha Kudugunta, Yacine Jernite, Mathias Jenny, Orhan Firat, Bonaventure F.~P. Dossou, Sakhile Dlamini, Nisansa de~Silva, Sakine {\c{C}}abuk~Ball{\i}, Stella Biderman, Alessia Battisti, Ahmed Baruwa, Ankur Bapna, Pallavi Baljekar, Israel~Abebe Azime, Ayodele Awokoya, Duygu Ataman, Orevaoghene Ahia, Oghenefego Ahia, Sweta Agrawal, and Mofetoluwa Adeyemi.
\newblock Quality at a glance: An audit of web-crawled multilingual datasets.
\newblock \emph{Transactions of the Association for Computational Linguistics}, 10:\penalty0 50--72, 2022.
\newblock \doi{10.1162/tacl_a_00447}.
\newblock URL \url{https://aclanthology.org/2022.tacl-1.4}.

\bibitem[Kudo \& Richardson(2018)Kudo and Richardson]{kudo-richardson-2018-sentencepiece}
Taku Kudo and John Richardson.
\newblock {S}entence{P}iece: A simple and language independent subword tokenizer and detokenizer for neural text processing.
\newblock In \emph{Proceedings of the 2018 Conference on Empirical Methods in Natural Language Processing: System Demonstrations}, pp.\  66--71, Brussels, Belgium, November 2018. Association for Computational Linguistics.
\newblock \doi{10.18653/v1/D18-2012}.
\newblock URL \url{https://aclanthology.org/D18-2012}.

\bibitem[Lample et~al.(2018)Lample, Conneau, Denoyer, and Ranzato]{lample2018unsupervised}
Guillaume Lample, Alexis Conneau, Ludovic Denoyer, and Marc'Aurelio Ranzato.
\newblock Unsupervised machine translation using monolingual corpora only.
\newblock In \emph{International Conference on Learning Representations}, 2018.
\newblock URL \url{https://openreview.net/forum?id=rkYTTf-AZ}.

\bibitem[Li et~al.(2018)Li, He, Ren, and Mao]{10.1007/978-3-030-05677-3_10}
Junyu Li, Ligang He, Shenyuan Ren, and Rui Mao.
\newblock Data fine-pruning: A simple way to accelerate neural network training.
\newblock In Feng Zhang, Jidong Zhai, Marc Snir, Hai Jin, Hironori Kasahara, and Mateo Valero (eds.), \emph{Network and Parallel Computing}, pp.\  114--125, Cham, 2018. Springer International Publishing.
\newblock ISBN 978-3-030-05677-3.

\bibitem[Liu et~al.(2020)Liu, Gu, Goyal, Li, Edunov, Ghazvininejad, Lewis, and Zettlemoyer]{liu-etal-2020-multilingual-denoising}
Yinhan Liu, Jiatao Gu, Naman Goyal, Xian Li, Sergey Edunov, Marjan Ghazvininejad, Mike Lewis, and Luke Zettlemoyer.
\newblock Multilingual denoising pre-training for neural machine translation.
\newblock \emph{Transactions of the Association for Computational Linguistics}, 8:\penalty0 726--742, 2020.
\newblock \doi{10.1162/tacl_a_00343}.
\newblock URL \url{https://aclanthology.org/2020.tacl-1.47}.

\bibitem[Longpre et~al.(2023)Longpre, Mahari, Chen, Obeng-Marnu, Sileo, Brannon, Muennighoff, Khazam, Kabbara, Perisetla, Wu, Shippole, Bollacker, Wu, Villa, Pentland, and Hooker]{longpre2023data}
Shayne Longpre, Robert Mahari, Anthony Chen, Naana Obeng-Marnu, Damien Sileo, William Brannon, Niklas Muennighoff, Nathan Khazam, Jad Kabbara, Kartik Perisetla, Xinyi Wu, Enrico Shippole, Kurt Bollacker, Tongshuang Wu, Luis Villa, Sandy Pentland, and Sara Hooker.
\newblock The data provenance initiative: A large scale audit of dataset licensing \& attribution in ai, 2023.

\bibitem[Marion et~al.(2023)Marion, Üstün, Pozzobon, Wang, Fadaee, and Hooker]{marion2023more}
Max Marion, Ahmet Üstün, Luiza Pozzobon, Alex Wang, Marzieh Fadaee, and Sara Hooker.
\newblock When less is more: Investigating data pruning for pretraining llms at scale, 2023.

\bibitem[Moore \& Lewis(2010)Moore and Lewis]{moore-lewis-2010-intelligent}
Robert~C. Moore and William Lewis.
\newblock Intelligent selection of language model training data.
\newblock In \emph{Proceedings of the {ACL} 2010 Conference Short Papers}, pp.\  220--224, Uppsala, Sweden, July 2010. Association for Computational Linguistics.
\newblock URL \url{https://aclanthology.org/P10-2041}.

\bibitem[Nair \& Hinton(2010)Nair and Hinton]{Nair2010RectifiedLU}
Vinod Nair and Geoffrey~E. Hinton.
\newblock Rectified linear units improve restricted boltzmann machines.
\newblock In \emph{International Conference on Machine Learning}, 2010.

\bibitem[Ott et~al.(2019)Ott, Edunov, Baevski, Fan, Gross, Ng, Grangier, and Auli]{ott2019fairseq}
Myle Ott, Sergey Edunov, Alexei Baevski, Angela Fan, Sam Gross, Nathan Ng, David Grangier, and Michael Auli.
\newblock fairseq: A fast, extensible toolkit for sequence modeling.
\newblock In \emph{Proceedings of NAACL-HLT 2019: Demonstrations}, 2019.

\bibitem[Papineni et~al.(2002)Papineni, Roukos, Ward, and Zhu]{papineni-etal-2002-bleu}
Kishore Papineni, Salim Roukos, Todd Ward, and Wei-Jing Zhu.
\newblock {B}leu: a method for automatic evaluation of machine translation.
\newblock In \emph{Proceedings of the 40th Annual Meeting of the Association for Computational Linguistics}, pp.\  311--318, Philadelphia, Pennsylvania, USA, July 2002. Association for Computational Linguistics.
\newblock \doi{10.3115/1073083.1073135}.
\newblock URL \url{https://aclanthology.org/P02-1040}.

\bibitem[Post(2018)]{post-2018-call}
Matt Post.
\newblock A call for clarity in reporting {BLEU} scores.
\newblock In \emph{Proceedings of the Third Conference on Machine Translation: Research Papers}, pp.\  186--191, Brussels, Belgium, October 2018. Association for Computational Linguistics.
\newblock \doi{10.18653/v1/W18-6319}.
\newblock URL \url{https://aclanthology.org/W18-6319}.

\bibitem[Raju et~al.(2021)Raju, Daruwalla, and Lipasti]{Raju2021AcceleratingDL}
Ravi Raju, Kyle Daruwalla, and Mikko~H. Lipasti.
\newblock Accelerating deep learning with dynamic data pruning.
\newblock \emph{ArXiv}, abs/2111.12621, 2021.

\bibitem[Ramesh et~al.(2022)Ramesh, Doddapaneni, Bheemaraj, Jobanputra, AK, Sharma, Sahoo, Diddee, J, Kakwani, Kumar, Pradeep, Nagaraj, Deepak, Raghavan, Kunchukuttan, Kumar, and Khapra]{ramesh-etal-2022-samanantar}
Gowtham Ramesh, Sumanth Doddapaneni, Aravinth Bheemaraj, Mayank Jobanputra, Raghavan AK, Ajitesh Sharma, Sujit Sahoo, Harshita Diddee, Mahalakshmi J, Divyanshu Kakwani, Navneet Kumar, Aswin Pradeep, Srihari Nagaraj, Kumar Deepak, Vivek Raghavan, Anoop Kunchukuttan, Pratyush Kumar, and Mitesh~Shantadevi Khapra.
\newblock Samanantar: The largest publicly available parallel corpora collection for 11 {I}ndic languages.
\newblock \emph{Transactions of the Association for Computational Linguistics}, 10:\penalty0 145--162, 2022.
\newblock \doi{10.1162/tacl_a_00452}.
\newblock URL \url{https://aclanthology.org/2022.tacl-1.9}.

\bibitem[Rei et~al.(2020)Rei, Stewart, Farinha, and Lavie]{rei-etal-2020-unbabels}
Ricardo Rei, Craig Stewart, Ana~C Farinha, and Alon Lavie.
\newblock Unbabel{'}s participation in the {WMT}20 metrics shared task.
\newblock In \emph{Proceedings of the Fifth Conference on Machine Translation}, pp.\  911--920, Online, November 2020. Association for Computational Linguistics.
\newblock URL \url{https://aclanthology.org/2020.wmt-1.101}.

\bibitem[Reid et~al.(2021)Reid, Hu, Neubig, and Matsuo]{reid-etal-2021-afromt}
Machel Reid, Junjie Hu, Graham Neubig, and Yutaka Matsuo.
\newblock {A}fro{MT}: Pretraining strategies and reproducible benchmarks for translation of 8 {A}frican languages.
\newblock In \emph{Proceedings of the 2021 Conference on Empirical Methods in Natural Language Processing}, pp.\  1306--1320, Online and Punta Cana, Dominican Republic, November 2021. Association for Computational Linguistics.
\newblock \doi{10.18653/v1/2021.emnlp-main.99}.
\newblock URL \url{https://aclanthology.org/2021.emnlp-main.99}.

\bibitem[Scao et~al.(2022)Scao, Fan, Akiki, Pavlick, Ilić, Hesslow, Castagné, Luccioni, Yvon, Gallé, Tow, Rush, Biderman, Webson, Ammanamanchi, Wang, Sagot, Muennighoff, del Moral, Ruwase, Bawden, Bekman, McMillan-Major, Beltagy, Nguyen, Saulnier, Tan, Suarez, Sanh, Laurençon, Jernite, Launay, Mitchell, Raffel, Gokaslan, Simhi, Soroa, Aji, Alfassy, Rogers, Nitzav, Xu, Mou, Emezue, Klamm, Leong, van Strien, Adelani, Radev, Ponferrada, Levkovizh, Kim, Natan, Toni, Dupont, Kruszewski, Pistilli, Elsahar, Benyamina, Tran, Yu, Abdulmumin, Johnson, Gonzalez-Dios, de~la Rosa, Chim, Dodge, Zhu, Chang, Frohberg, Tobing, Bhattacharjee, Almubarak, Chen, Lo, Werra, Weber, Phan, allal, Tanguy, Dey, Muñoz, Masoud, Grandury, Šaško, Huang, Coavoux, Singh, Jiang, Vu, Jauhar, Ghaleb, Subramani, Kassner, Khamis, Nguyen, Espejel, de~Gibert, Villegas, Henderson, Colombo, Amuok, Lhoest, Harliman, Bommasani, López, Ribeiro, Osei, Pyysalo, Nagel, Bose, Muhammad, Sharma, Longpre, Nikpoor, Silberberg, Pai, Zink, Torrent,
  Schick, Thrush, Danchev, Nikoulina, Laippala, Lepercq, Prabhu, Alyafeai, Talat, Raja, Heinzerling, Si, Taşar, Salesky, Mielke, Lee, Sharma, Santilli, Chaffin, Stiegler, Datta, Szczechla, Chhablani, Wang, Pandey, Strobelt, Fries, Rozen, Gao, Sutawika, Bari, Al-shaibani, Manica, Nayak, Teehan, Albanie, Shen, Ben-David, Bach, Kim, Bers, Fevry, Neeraj, Thakker, Raunak, Tang, Yong, Sun, Brody, Uri, Tojarieh, Roberts, Chung, Tae, Phang, Press, Li, Narayanan, Bourfoune, Casper, Rasley, Ryabinin, Mishra, Zhang, Shoeybi, Peyrounette, Patry, Tazi, Sanseviero, von Platen, Cornette, Lavallée, Lacroix, Rajbhandari, Gandhi, Smith, Requena, Patil, Dettmers, Baruwa, Singh, Cheveleva, Ligozat, Subramonian, Névéol, Lovering, Garrette, Tunuguntla, Reiter, Taktasheva, Voloshina, Bogdanov, Winata, Schoelkopf, Kalo, Novikova, Forde, Clive, Kasai, Kawamura, Hazan, Carpuat, Clinciu, Kim, Cheng, Serikov, Antverg, van~der Wal, Zhang, Zhang, Gehrmann, Mirkin, Pais, Shavrina, Scialom, Yun, Limisiewicz, Rieser, Protasov, Mikhailov,
  Pruksachatkun, Belinkov, Bamberger, Kasner, Rueda, Pestana, Feizpour, Khan, Faranak, Santos, Hevia, Unldreaj, Aghagol, Abdollahi, Tammour, HajiHosseini, Behroozi, Ajibade, Saxena, Ferrandis, McDuff, Contractor, Lansky, David, Kiela, Nguyen, Tan, Baylor, Ozoani, Mirza, Ononiwu, Rezanejad, Jones, Bhattacharya, Solaiman, Sedenko, Nejadgholi, Passmore, Seltzer, Sanz, Dutra, Samagaio, Elbadri, Mieskes, Gerchick, Akinlolu, McKenna, Qiu, Ghauri, Burynok, Abrar, Rajani, Elkott, Fahmy, Samuel, An, Kromann, Hao, Alizadeh, Shubber, Wang, Roy, Viguier, Le, Oyebade, Le, Yang, Nguyen, Kashyap, Palasciano, Callahan, Shukla, Miranda-Escalada, Singh, Beilharz, Wang, Brito, Zhou, Jain, Xu, Fourrier, Periñán, Molano, Yu, Manjavacas, Barth, Fuhrimann, Altay, Bayrak, Burns, Vrabec, Bello, Dash, Kang, Giorgi, Golde, Posada, Sivaraman, Bulchandani, Liu, Shinzato, de~Bykhovetz, Takeuchi, Pàmies, Castillo, Nezhurina, Sänger, Samwald, Cullan, Weinberg, Wolf, Mihaljcic, Liu, Freidank, Kang, Seelam, Dahlberg, Broad, Muellner,
  Fung, Haller, Chandrasekhar, Eisenberg, Martin, Canalli, Su, Su, Cahyawijaya, Garda, Deshmukh, Mishra, Kiblawi, Ott, Sang-aroonsiri, Kumar, Schweter, Bharati, Laud, Gigant, Kainuma, Kusa, Labrak, Bajaj, Venkatraman, Xu, Xu, Xu, Tan, Xie, Ye, Bras, Belkada, and Wolf]{bloom}
Teven~Le Scao, Angela Fan, Christopher Akiki, Ellie Pavlick, Suzana Ilić, Daniel Hesslow, Roman Castagné, Alexandra~Sasha Luccioni, François Yvon, Matthias Gallé, Jonathan Tow, Alexander~M. Rush, Stella Biderman, Albert Webson, Pawan~Sasanka Ammanamanchi, Thomas Wang, Benoît Sagot, Niklas Muennighoff, Albert~Villanova del Moral, Olatunji Ruwase, Rachel Bawden, Stas Bekman, Angelina McMillan-Major, Iz~Beltagy, Huu Nguyen, Lucile Saulnier, Samson Tan, Pedro~Ortiz Suarez, Victor Sanh, Hugo Laurençon, Yacine Jernite, Julien Launay, Margaret Mitchell, Colin Raffel, Aaron Gokaslan, Adi Simhi, Aitor Soroa, Alham~Fikri Aji, Amit Alfassy, Anna Rogers, Ariel~Kreisberg Nitzav, Canwen Xu, Chenghao Mou, Chris Emezue, Christopher Klamm, Colin Leong, Daniel van Strien, David~Ifeoluwa Adelani, Dragomir Radev, Eduardo~González Ponferrada, Efrat Levkovizh, Ethan Kim, Eyal~Bar Natan, Francesco~De Toni, Gérard Dupont, Germán Kruszewski, Giada Pistilli, Hady Elsahar, Hamza Benyamina, Hieu Tran, Ian Yu, Idris Abdulmumin,
  Isaac Johnson, Itziar Gonzalez-Dios, Javier de~la Rosa, Jenny Chim, Jesse Dodge, Jian Zhu, Jonathan Chang, Jörg Frohberg, Joseph Tobing, Joydeep Bhattacharjee, Khalid Almubarak, Kimbo Chen, Kyle Lo, Leandro~Von Werra, Leon Weber, Long Phan, Loubna~Ben allal, Ludovic Tanguy, Manan Dey, Manuel~Romero Muñoz, Maraim Masoud, María Grandury, Mario Šaško, Max Huang, Maximin Coavoux, Mayank Singh, Mike Tian-Jian Jiang, Minh~Chien Vu, Mohammad~A. Jauhar, Mustafa Ghaleb, Nishant Subramani, Nora Kassner, Nurulaqilla Khamis, Olivier Nguyen, Omar Espejel, Ona de~Gibert, Paulo Villegas, Peter Henderson, Pierre Colombo, Priscilla Amuok, Quentin Lhoest, Rheza Harliman, Rishi Bommasani, Roberto~Luis López, Rui Ribeiro, Salomey Osei, Sampo Pyysalo, Sebastian Nagel, Shamik Bose, Shamsuddeen~Hassan Muhammad, Shanya Sharma, Shayne Longpre, Somaieh Nikpoor, Stanislav Silberberg, Suhas Pai, Sydney Zink, Tiago~Timponi Torrent, Timo Schick, Tristan Thrush, Valentin Danchev, Vassilina Nikoulina, Veronika Laippala, Violette
  Lepercq, Vrinda Prabhu, Zaid Alyafeai, Zeerak Talat, Arun Raja, Benjamin Heinzerling, Chenglei Si, Davut~Emre Taşar, Elizabeth Salesky, Sabrina~J. Mielke, Wilson~Y. Lee, Abheesht Sharma, Andrea Santilli, Antoine Chaffin, Arnaud Stiegler, Debajyoti Datta, Eliza Szczechla, Gunjan Chhablani, Han Wang, Harshit Pandey, Hendrik Strobelt, Jason~Alan Fries, Jos Rozen, Leo Gao, Lintang Sutawika, M~Saiful Bari, Maged~S. Al-shaibani, Matteo Manica, Nihal Nayak, Ryan Teehan, Samuel Albanie, Sheng Shen, Srulik Ben-David, Stephen~H. Bach, Taewoon Kim, Tali Bers, Thibault Fevry, Trishala Neeraj, Urmish Thakker, Vikas Raunak, Xiangru Tang, Zheng-Xin Yong, Zhiqing Sun, Shaked Brody, Yallow Uri, Hadar Tojarieh, Adam Roberts, Hyung~Won Chung, Jaesung Tae, Jason Phang, Ofir Press, Conglong Li, Deepak Narayanan, Hatim Bourfoune, Jared Casper, Jeff Rasley, Max Ryabinin, Mayank Mishra, Minjia Zhang, Mohammad Shoeybi, Myriam Peyrounette, Nicolas Patry, Nouamane Tazi, Omar Sanseviero, Patrick von Platen, Pierre Cornette,
  Pierre~François Lavallée, Rémi Lacroix, Samyam Rajbhandari, Sanchit Gandhi, Shaden Smith, Stéphane Requena, Suraj Patil, Tim Dettmers, Ahmed Baruwa, Amanpreet Singh, Anastasia Cheveleva, Anne-Laure Ligozat, Arjun Subramonian, Aurélie Névéol, Charles Lovering, Dan Garrette, Deepak Tunuguntla, Ehud Reiter, Ekaterina Taktasheva, Ekaterina Voloshina, Eli Bogdanov, Genta~Indra Winata, Hailey Schoelkopf, Jan-Christoph Kalo, Jekaterina Novikova, Jessica~Zosa Forde, Jordan Clive, Jungo Kasai, Ken Kawamura, Liam Hazan, Marine Carpuat, Miruna Clinciu, Najoung Kim, Newton Cheng, Oleg Serikov, Omer Antverg, Oskar van~der Wal, Rui Zhang, Ruochen Zhang, Sebastian Gehrmann, Shachar Mirkin, Shani Pais, Tatiana Shavrina, Thomas Scialom, Tian Yun, Tomasz Limisiewicz, Verena Rieser, Vitaly Protasov, Vladislav Mikhailov, Yada Pruksachatkun, Yonatan Belinkov, Zachary Bamberger, Zdeněk Kasner, Alice Rueda, Amanda Pestana, Amir Feizpour, Ammar Khan, Amy Faranak, Ana Santos, Anthony Hevia, Antigona Unldreaj, Arash Aghagol,
  Arezoo Abdollahi, Aycha Tammour, Azadeh HajiHosseini, Bahareh Behroozi, Benjamin Ajibade, Bharat Saxena, Carlos~Muñoz Ferrandis, Daniel McDuff, Danish Contractor, David Lansky, Davis David, Douwe Kiela, Duong~A. Nguyen, Edward Tan, Emi Baylor, Ezinwanne Ozoani, Fatima Mirza, Frankline Ononiwu, Habib Rezanejad, Hessie Jones, Indrani Bhattacharya, Irene Solaiman, Irina Sedenko, Isar Nejadgholi, Jesse Passmore, Josh Seltzer, Julio~Bonis Sanz, Livia Dutra, Mairon Samagaio, Maraim Elbadri, Margot Mieskes, Marissa Gerchick, Martha Akinlolu, Michael McKenna, Mike Qiu, Muhammed Ghauri, Mykola Burynok, Nafis Abrar, Nazneen Rajani, Nour Elkott, Nour Fahmy, Olanrewaju Samuel, Ran An, Rasmus Kromann, Ryan Hao, Samira Alizadeh, Sarmad Shubber, Silas Wang, Sourav Roy, Sylvain Viguier, Thanh Le, Tobi Oyebade, Trieu Le, Yoyo Yang, Zach Nguyen, Abhinav~Ramesh Kashyap, Alfredo Palasciano, Alison Callahan, Anima Shukla, Antonio Miranda-Escalada, Ayush Singh, Benjamin Beilharz, Bo~Wang, Caio Brito, Chenxi Zhou, Chirag Jain,
  Chuxin Xu, Clémentine Fourrier, Daniel~León Periñán, Daniel Molano, Dian Yu, Enrique Manjavacas, Fabio Barth, Florian Fuhrimann, Gabriel Altay, Giyaseddin Bayrak, Gully Burns, Helena~U. Vrabec, Imane Bello, Ishani Dash, Jihyun Kang, John Giorgi, Jonas Golde, Jose~David Posada, Karthik~Rangasai Sivaraman, Lokesh Bulchandani, Lu~Liu, Luisa Shinzato, Madeleine~Hahn de~Bykhovetz, Maiko Takeuchi, Marc Pàmies, Maria~A Castillo, Marianna Nezhurina, Mario Sänger, Matthias Samwald, Michael Cullan, Michael Weinberg, Michiel~De Wolf, Mina Mihaljcic, Minna Liu, Moritz Freidank, Myungsun Kang, Natasha Seelam, Nathan Dahlberg, Nicholas~Michio Broad, Nikolaus Muellner, Pascale Fung, Patrick Haller, Ramya Chandrasekhar, Renata Eisenberg, Robert Martin, Rodrigo Canalli, Rosaline Su, Ruisi Su, Samuel Cahyawijaya, Samuele Garda, Shlok~S Deshmukh, Shubhanshu Mishra, Sid Kiblawi, Simon Ott, Sinee Sang-aroonsiri, Srishti Kumar, Stefan Schweter, Sushil Bharati, Tanmay Laud, Théo Gigant, Tomoya Kainuma, Wojciech Kusa, Yanis
  Labrak, Yash~Shailesh Bajaj, Yash Venkatraman, Yifan Xu, Yingxin Xu, Yu~Xu, Zhe Tan, Zhongli Xie, Zifan Ye, Mathilde Bras, Younes Belkada, and Thomas Wolf.
\newblock Bloom: A 176b-parameter open-access multilingual language model.
\newblock \emph{arXiv preprint arXiv: 2211.05100}, 2022.
\newblock URL \url{http://arxiv.org/abs/2211.05100}.

\bibitem[Schwenk(2018)]{schwenk-2018-filtering}
Holger Schwenk.
\newblock Filtering and mining parallel data in a joint multilingual space.
\newblock In \emph{Proceedings of the 56th Annual Meeting of the Association for Computational Linguistics (Volume 2: Short Papers)}, pp.\  228--234, Melbourne, Australia, July 2018. Association for Computational Linguistics.
\newblock \doi{10.18653/v1/P18-2037}.
\newblock URL \url{https://aclanthology.org/P18-2037}.

\bibitem[Schwenk et~al.(2021{\natexlab{a}})Schwenk, Chaudhary, Sun, Gong, and Guzm{\'a}n]{schwenk-etal-2021-wikimatrix}
Holger Schwenk, Vishrav Chaudhary, Shuo Sun, Hongyu Gong, and Francisco Guzm{\'a}n.
\newblock {W}iki{M}atrix: Mining 135{M} parallel sentences in 1620 language pairs from {W}ikipedia.
\newblock In \emph{Proceedings of the 16th Conference of the European Chapter of the Association for Computational Linguistics: Main Volume}, pp.\  1351--1361, Online, April 2021{\natexlab{a}}. Association for Computational Linguistics.
\newblock \doi{10.18653/v1/2021.eacl-main.115}.
\newblock URL \url{https://aclanthology.org/2021.eacl-main.115}.

\bibitem[Schwenk et~al.(2021{\natexlab{b}})Schwenk, Wenzek, Edunov, Grave, Joulin, and Fan]{schwenk-etal-2021-ccmatrix}
Holger Schwenk, Guillaume Wenzek, Sergey Edunov, Edouard Grave, Armand Joulin, and Angela Fan.
\newblock {CCM}atrix: Mining billions of high-quality parallel sentences on the web.
\newblock In \emph{Proceedings of the 59th Annual Meeting of the Association for Computational Linguistics and the 11th International Joint Conference on Natural Language Processing (Volume 1: Long Papers)}, pp.\  6490--6500, Online, August 2021{\natexlab{b}}. Association for Computational Linguistics.
\newblock \doi{10.18653/v1/2021.acl-long.507}.
\newblock URL \url{https://aclanthology.org/2021.acl-long.507}.

\bibitem[Siddiqui et~al.(2022)Siddiqui, Rajkumar, Maharaj, Krueger, and Hooker]{siddiqui2022metadata}
Shoaib~Ahmed Siddiqui, Nitarshan Rajkumar, Tegan Maharaj, David Krueger, and Sara Hooker.
\newblock Metadata archaeology: Unearthing data subsets by leveraging training dynamics, 2022.

\bibitem[Sorscher et~al.(2022)Sorscher, Geirhos, Shekhar, Ganguli, and Morcos]{sorscher2023neural}
Ben Sorscher, Robert Geirhos, Shashank Shekhar, Surya Ganguli, and Ari~S. Morcos.
\newblock Beyond neural scaling laws: beating power law scaling via data pruning.
\newblock In Alice~H. Oh, Alekh Agarwal, Danielle Belgrave, and Kyunghyun Cho (eds.), \emph{Advances in Neural Information Processing Systems}, 2022.
\newblock URL \url{https://openreview.net/forum?id=UmvSlP-PyV}.

\bibitem[Srivastava et~al.(2014)Srivastava, Hinton, Krizhevsky, Sutskever, and Salakhutdinov]{dropout}
Nitish Srivastava, Geoffrey Hinton, Alex Krizhevsky, Ilya Sutskever, and Ruslan Salakhutdinov.
\newblock Dropout: A simple way to prevent neural networks from overfitting.
\newblock \emph{J. Mach. Learn. Res.}, 15\penalty0 (1):\penalty0 1929–1958, jan 2014.
\newblock ISSN 1532-4435.

\bibitem[Steingr{\'\i}msson et~al.(2021)Steingr{\'\i}msson, Lohar, Loftsson, and Way]{steingrimsson-etal-2021-effective}
Stein{\th}{\'o}r Steingr{\'\i}msson, Pintu Lohar, Hrafn Loftsson, and Andy Way.
\newblock Effective bitext extraction from comparable corpora using a combination of three different approaches.
\newblock In \emph{Proceedings of the 14th Workshop on Building and Using Comparable Corpora (BUCC 2021)}, pp.\  8--17, Online (Virtual Mode), September 2021. INCOMA Ltd.
\newblock URL \url{https://aclanthology.org/2021.bucc-1.3}.

\bibitem[Swayamdipta et~al.(2020)Swayamdipta, Schwartz, Lourie, Wang, Hajishirzi, Smith, and Choi]{swayamdipta-etal-2020-dataset}
Swabha Swayamdipta, Roy Schwartz, Nicholas Lourie, Yizhong Wang, Hannaneh Hajishirzi, Noah~A. Smith, and Yejin Choi.
\newblock Dataset cartography: Mapping and diagnosing datasets with training dynamics.
\newblock In \emph{Proceedings of the 2020 Conference on Empirical Methods in Natural Language Processing (EMNLP)}, pp.\  9275--9293, Online, November 2020. Association for Computational Linguistics.
\newblock \doi{10.18653/v1/2020.emnlp-main.746}.
\newblock URL \url{https://aclanthology.org/2020.emnlp-main.746}.

\bibitem[Szegedy et~al.(2016)Szegedy, Vanhoucke, Ioffe, Shlens, and Wojna]{labelsmoothing}
Christian Szegedy, Vincent Vanhoucke, Sergey Ioffe, Jon Shlens, and Zbigniew Wojna.
\newblock Rethinking the inception architecture for computer vision.
\newblock In \emph{2016 IEEE Conference on Computer Vision and Pattern Recognition (CVPR)}, pp.\  2818--2826, 2016.
\newblock \doi{10.1109/CVPR.2016.308}.

\bibitem[Thompson \& Koehn(2019)Thompson and Koehn]{thompson-koehn-2019-vecalign}
Brian Thompson and Philipp Koehn.
\newblock {V}ecalign: Improved sentence alignment in linear time and space.
\newblock In \emph{Proceedings of the 2019 Conference on Empirical Methods in Natural Language Processing and the 9th International Joint Conference on Natural Language Processing (EMNLP-IJCNLP)}, pp.\  1342--1348, Hong Kong, China, November 2019. Association for Computational Linguistics.
\newblock \doi{10.18653/v1/D19-1136}.
\newblock URL \url{https://aclanthology.org/D19-1136}.

\bibitem[van~der Wees et~al.(2017)van~der Wees, Bisazza, and Monz]{van-der-wees-etal-2017-dynamic}
Marlies van~der Wees, Arianna Bisazza, and Christof Monz.
\newblock Dynamic data selection for neural machine translation.
\newblock In \emph{Proceedings of the 2017 Conference on Empirical Methods in Natural Language Processing}, pp.\  1400--1410, Copenhagen, Denmark, September 2017. Association for Computational Linguistics.
\newblock \doi{10.18653/v1/D17-1147}.
\newblock URL \url{https://aclanthology.org/D17-1147}.

\bibitem[Vaswani et~al.(2017)Vaswani, Shazeer, Parmar, Uszkoreit, Jones, Gomez, Kaiser, and Polosukhin]{DBLP:journals/corr/VaswaniSPUJGKP17}
Ashish Vaswani, Noam Shazeer, Niki Parmar, Jakob Uszkoreit, Llion Jones, Aidan~N. Gomez, Lukasz Kaiser, and Illia Polosukhin.
\newblock Attention is all you need.
\newblock \emph{CoRR}, abs/1706.03762, 2017.
\newblock URL \url{http://arxiv.org/abs/1706.03762}.

\bibitem[Vegi et~al.(2022)Vegi, J, Paul, Mishra, Banjare, K~R, and Viswanathan]{vegi-etal-2022-webcrawl}
Pavanpankaj Vegi, Sivabhavani J, Biswajit Paul, Abhinav Mishra, Prashant Banjare, Prasanna K~R, and Chitra Viswanathan.
\newblock {W}eb{C}rawl {A}frican : A multilingual parallel corpora for {A}frican languages.
\newblock In \emph{Proceedings of the Seventh Conference on Machine Translation (WMT)}, pp.\  1076--1089, Abu Dhabi, United Arab Emirates (Hybrid), December 2022. Association for Computational Linguistics.
\newblock URL \url{https://aclanthology.org/2022.wmt-1.105}.

\bibitem[Wang et~al.(2018)Wang, Watanabe, Hughes, Nakagawa, and Chelba]{wang-etal-2018-denoising}
Wei Wang, Taro Watanabe, Macduff Hughes, Tetsuji Nakagawa, and Ciprian Chelba.
\newblock Denoising neural machine translation training with trusted data and online data selection.
\newblock In \emph{Proceedings of the Third Conference on Machine Translation: Research Papers}, pp.\  133--143, Brussels, Belgium, October 2018. Association for Computational Linguistics.
\newblock \doi{10.18653/v1/W18-6314}.
\newblock URL \url{https://aclanthology.org/W18-6314}.

\bibitem[Xu et~al.(2023)Xu, Huang, JIANG, Cao, Yang, Wang, and Yang]{xu2023better}
Jiarong Xu, Renhong Huang, XIN JIANG, Yuxuan Cao, Carl Yang, Chunping Wang, and Yang Yang.
\newblock Better with less: A data-active perspective on pre-training graph neural networks.
\newblock In \emph{Thirty-seventh Conference on Neural Information Processing Systems}, 2023.
\newblock URL \url{https://openreview.net/forum?id=m2WR1yJ8N9}.

\bibitem[Zhang et~al.(2020)Zhang, Nagesh, and Knight]{zhang-etal-2020-parallel-corpus}
Boliang Zhang, Ajay Nagesh, and Kevin Knight.
\newblock Parallel corpus filtering via pre-trained language models.
\newblock In \emph{Proceedings of the 58th Annual Meeting of the Association for Computational Linguistics}, pp.\  8545--8554, Online, July 2020. Association for Computational Linguistics.
\newblock \doi{10.18653/v1/2020.acl-main.756}.
\newblock URL \url{https://aclanthology.org/2020.acl-main.756}.

\end{thebibliography}


\newpage
\appendix

\section{Dataset Analysis}

\begin{figure*}[ht]
  \centering
  \includegraphics[width=\textwidth]{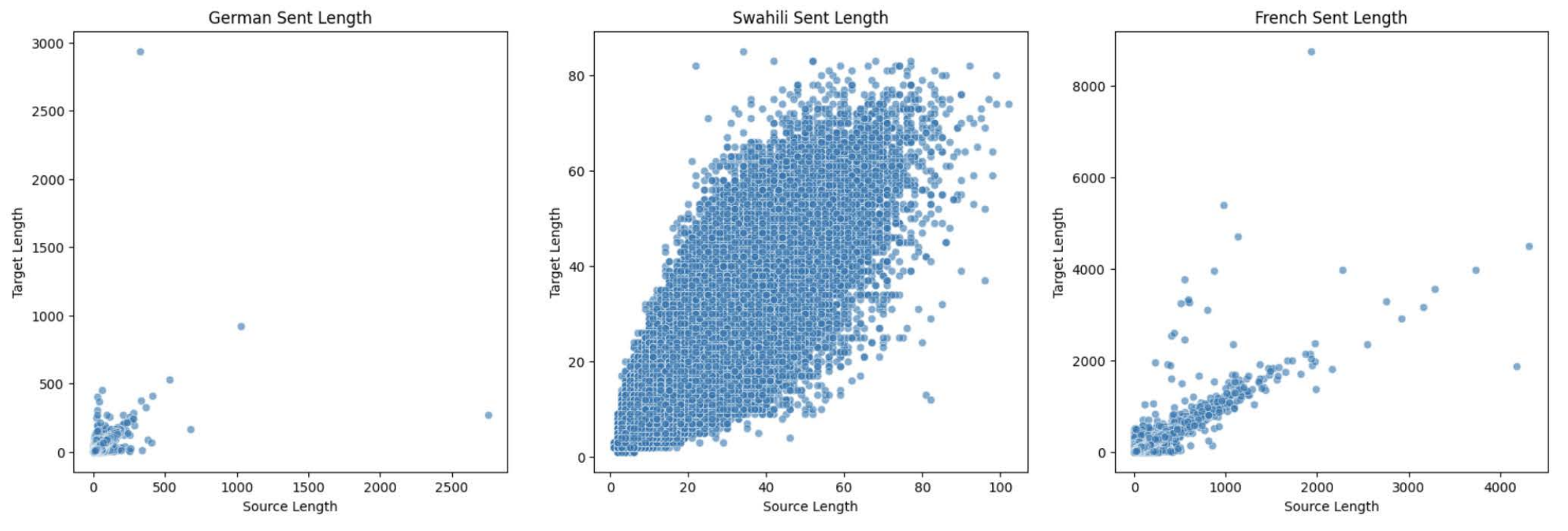}
  \caption{Source (English) sentence length vs Target (German, Swahili, French) sentence length. German and French show high variance while Swahili shows low variance.}
  \label{fig:Length}
\end{figure*}

\begin{figure*}[h!]
  \centering
  \includegraphics[width=\textwidth]{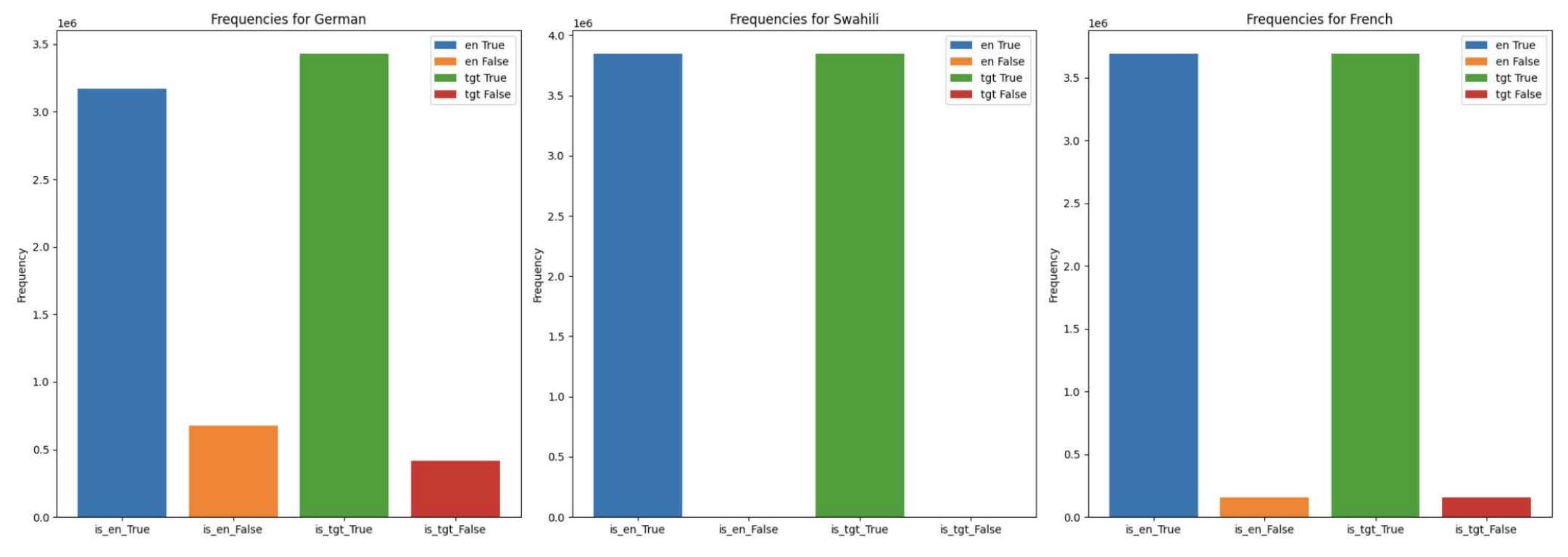}
  \caption{English sentences Language ID (en) vs Target (German, Swahili, French) sentence length (tgt). Swahili set does not contain any foreign languages while German and French seem to contain other languages.}
  \label{fig:ID}
\end{figure*}

\Cref{fig:Length} and \Cref{fig:ID} provide analysis of the 3.8M dataset for German, Swahili and French. Both German and French exhibit outliers in the sentence length with some sentences being 500 words or longer while Swahili seems to be filtered by sentence length as no sentence has more than 250 words. See \Cref{fig:Length}. We also see that foreign languages were removed from the Swahili dataset as both source and target sentences do not contain any other language whereas both German and French contain foreign language. 

\section{Other evaluation metrics and significance tests}

We ran additional experiments to calculate chrF++ and COMET scores and report the statistical significance results with paired bootstrap resampling \citep{koehn-2004-statistical} using sacreBLEU and COMET libraries with default configurations for the respective metrics below for all our experiments with on the 3.8M data scale. We find that CAT-DIFF demonstrates competitive or superior performance compared to other methods on en-de and en-fr pairs across most test sets. Bolded numbers in \Cref{tab:swahili,tab:german,tab:french} show that with 1000 resamples, CAT-DIFF outperforms the other technique 95\% of the time thus CAT-DIFF is statistically significance at 95\% confidence level. These results highlight the effectiveness of our proposed method.

\begin{table*}[t]
\small
\centering
\resizebox{0.9\textwidth}{!}{
\begin{tabular}{lcccccc}
\toprule
\multicolumn{1}{l}{{\textit{Swahili}}}                       & \multicolumn{3}{c}{\textit{FLORES}}                                                            & \multicolumn{3}{c}{\textit{MAFAND-MT}}                                      \\ \cmidrule{2-4} \cmidrule{5-7}
\multicolumn{1}{l}{\textbf{}}              & \textit{90\%}        & \textit{70\%}        & \multicolumn{1}{c}{\textit{50\%}}       & \textit{90\%}        & \textit{70\%}        & \textit{50\%}        \\ \midrule
\multicolumn{7}{c}{\textit{\text{BLEU}}} \\ \midrule
\multicolumn{1}{l|}{\text{CAT DIFF(1,5)}} & 25.0 ± 1.0           & 29.7 ± 1.0           & \multicolumn{1}{r|}{30.3 ± 1.1}          & 21.0 ± 0.8           & 25.0 ± 0.8           & 25.3 ± 0.9           \\
\multicolumn{1}{l|}{\text{Random}}        & \textbf{15.7 ± 0.8}  & \textbf{26.7 ± 0.9}  & \multicolumn{1}{r|}{28.7 ± 1.0}          & \textbf{14.4 ± 0.6}  & \textbf{22.0 ± 0.8}  & \textbf{23.6 ± 0.8}  \\
\multicolumn{1}{l|}{\text{LaBSE}}         & 29.2 ± 1.1  & 30.7 ± 1.0  & \multicolumn{1}{r|}{31.6 ± 1.0} & 24.9 ± 0.9  & 26.2 ± 0.9  & 26.2 ± 0.9  \\
\multicolumn{1}{l|}{\text{LASER}}         & \textbf{0.4 ± 0.1}   & \textbf{1.4 ± 0.3}   & \multicolumn{1}{r|}{\textbf{1.5 ± 0.3}}  & \textbf{2.1 ± 0.3}   & \textbf{3.2 ± 0.4}   & \textbf{3.6 ± 0.4}   \\
\multicolumn{1}{l|}{\text{COMET}}         & \textbf{15.3 ± 0.9}  & \textbf{27.6 ± 1.0}  & \multicolumn{1}{r|}{29.8 ± 1.0}          & \textbf{16.8 ± 0.8}  & \textbf{24.5 ± 0.8}  & 25.4 ± 0.9           \\
\multicolumn{1}{l|}{\text{BLOOM 560m}}    & \textbf{17.7 ± 0.9}  & \textbf{27.0 ± 0.9}  & \multicolumn{1}{r|}{\textbf{28.7 ± 1.0}} & \textbf{15.0 ± 0.7}  & \textbf{22.1 ± 0.8}  & \textbf{24.2 ± 0.8}  \\
\multicolumn{1}{l|}{\text{BLOOM 1b}}      & \textbf{17.2 ± 0.9}  & \textbf{26.5 ± 1.0}  & \multicolumn{1}{r|}{\textbf{29.4 ± 1.1}} & \textbf{14.5 ± 0.6}  & \textbf{21.9 ± 0.8}  & \textbf{24.2 ± 0.8}  \\
\multicolumn{1}{l|}{\text{BLOOM 1.7b}}    & \textbf{14.5 ± 0.8}  & \textbf{26.4 ± 1.0}  & \multicolumn{1}{r|}{\textbf{28.1 ± 1.0}} & \textbf{12.5 ± 0.6}  & \textbf{21.7 ± 0.8}  & \textbf{23.5 ± 0.8}  \\
\multicolumn{1}{l}{\text{Full}}          & 30.9 ± 1.0  & 30.9 ± 1.0  & \multicolumn{1}{r|}{30.9 ± 1.0}          & 27.5 ± 1.0  & 27.5 ± 1.0  & 27.5 ± 1.0  \\ \midrule
\multicolumn{7}{c}{\textit{\text{chrF++}}} \\ \midrule
\multicolumn{1}{l|}{\text{CAT DIFF(1,5)}} & 52.2 ± 0.7           & 56.0 ± 0.7           & \multicolumn{1}{r|}{56.4 ± 0.8}          & 47.2 ± 0.7           & 50.9 ± 0.7           & 51.0 ± 0.7           \\
\multicolumn{1}{l|}{\text{Random}}        & \textbf{42.7 ± 0.7}  & \textbf{53.5 ± 0.7}  & \multicolumn{1}{r|}{\textbf{55.0 ± 0.7}} & \textbf{39.8 ± 0.6}  & \textbf{48.5 ± 0.6}  & \textbf{49.7 ± 0.7}  \\
\multicolumn{1}{l|}{\text{LaBSE}}         & 55.9 ± 0.8  & 57.0 ± 0.7  & \multicolumn{1}{r|}{57.5 ± 0.7} & 51.1 ± 0.6  & 52.3 ± 0.7  & 52.3 ± 0.7  \\
\multicolumn{1}{l|}{\text{LASER}}         & \textbf{8.2 ± 0.3}   & \textbf{13.0 ± 0.3}  & \multicolumn{1}{r|}{\textbf{13.2 ± 0.3}} & \textbf{12.5 ± 0.3}  & \textbf{14.6 ± 0.4}  & \textbf{15.5 ± 0.4}  \\
\multicolumn{1}{l|}{\text{COMET}}         & \textbf{42.2 ± 0.7}  & \textbf{54.2 ± 0.7}  & \multicolumn{1}{r|}{56.2 ± 0.7}          & \textbf{42.3 ± 0.7}  & \textbf{50.6 ± 0.7}  & \textbf{51.3 ± 0.7}  \\
\multicolumn{1}{l|}{\text{BLOOM 560m}}    & \textbf{44.9 ± 0.7}  & \textbf{53.6 ± 0.7}  & \multicolumn{1}{r|}{\textbf{55.3 ± 0.8}} & \textbf{41.2 ± 0.6}  & \textbf{48.6 ± 0.6}  & \textbf{50.1 ± 0.7}  \\
\multicolumn{1}{l|}{\text{BLOOM 1b}}      & \textbf{44.1 ± 0.7}  & \textbf{53.2 ± 0.7}  & \multicolumn{1}{r|}{\textbf{55.5 ± 0.8}} & \textbf{40.2 ± 0.6}  & \textbf{48.2 ± 0.7}  & \textbf{50.3 ± 0.7}  \\
\multicolumn{1}{l|}{\text{BLOOM 1.7b}}    & \textbf{40.7 ± 0.7}  & \textbf{53.0 ± 0.8}  & \multicolumn{1}{r|}{\textbf{54.7 ± 0.8}} & \textbf{37.4 ± 0.5}  & \textbf{48.1 ± 0.7}  & \textbf{49.6 ± 0.7}  \\
\multicolumn{1}{l|}{\text{Full}}          & 56.8 ± 0.7  & 56.8 ± 0.7  & \multicolumn{1}{r|}{56.8 ± 0.7}          & 52.9 ± 0.7  & 52.9 ± 0.7  & 52.9 ± 0.7  \\ \midrule
\multicolumn{7}{c}{\textit{\text{COMET}}} \\ \midrule
\multicolumn{1}{l|}{\text{CAT DIFF(1,5)}} & 75.81                & 79.25                & \multicolumn{1}{c|}{79.49}               & 76.38                & 80.18                & 80.45                \\
\multicolumn{1}{l|}{\text{Random}}        & \textbf{68.97}       & \textbf{77.66}       & \multicolumn{1}{c|}{\textbf{79.05}}      & \textbf{71.30}        & \textbf{78.39}       & \textbf{79.66}       \\
\multicolumn{1}{l|}{\text{LaBSE}}         & 78.29                & 79.72                & \multicolumn{1}{c|}{80.36}               & 80.30                 & 81.65                & 81.75                \\
\multicolumn{1}{l|}{\text{LASER}}         & \textbf{30.82}       & \textbf{41.91}       & \multicolumn{1}{c|}{\textbf{44.56}}      & \textbf{40.43}       & \textbf{45.88}       & \textbf{48.23}       \\
\multicolumn{1}{l|}{\text{COMET}}         & \textbf{71.42}       & 79.39                & \multicolumn{1}{c|}{80.43}               & \textbf{75.70}        & 81.33                & 81.18                \\
\multicolumn{1}{l|}{\text{BLOOM 560m}}    & \textbf{70.19}       & \textbf{77.69}       & \multicolumn{1}{c|}{79.10}                & \textbf{70.91}       & \textbf{78.74}       & 80.27                \\
\multicolumn{1}{l|}{\text{BLOOM 1b}}      & \textbf{69.31}       & \textbf{77.12}       & \multicolumn{1}{c|}{\textbf{78.95}}      & \textbf{69.84}       & \textbf{78.43}       & \textbf{79.90}        \\
\multicolumn{1}{l|}{\text{BLOOM 1.7b}}    & \textbf{68.15}       & \textbf{77.34}       & \multicolumn{1}{c|}{\textbf{78.56}}      & \textbf{71.85}       & \textbf{78.33}       & \textbf{79.48}       \\
\multicolumn{1}{l|}{\text{Full}}          & 80.01                & 80.01                & \multicolumn{1}{c|}{80.01}               & 81.81                & 81.81                & 81.81   \\
\toprule
\end{tabular}}
\caption{Significance test for Swahili with the null hypothesis: mean translation score of CAT-DIFF is equal to the mean of the other technique. A bolded entry signifies that CAT-DIFF is significantly better (p-value $< 0.05$) than the method used for the corresponding value. For each dataset, we show results for different prune levels (90\%, 70\%, 50\%) and metrics (BLEU, chrF++, COMET). Here "Full" means the value obtained when using all the training data. 
}
\label{tab:swahili}
\end{table*}

\begin{table*}[t]
\small
\centering
\resizebox{0.9\textwidth}{!}{
\begin{tabular}{lcccccc}
\toprule
\multicolumn{1}{l}{{\textit{German}}}                       & \multicolumn{3}{c}{\textit{FLORES}}                                                   & \multicolumn{3}{c}{\textit{WMT18}}                                 \\ \cmidrule{2-7}
\multicolumn{1}{l}{\textbf{}}              & \textit{90\%}        & \textit{70\%}        & \multicolumn{1}{c}{\textit{50\%}}       & \textit{90\%}        & \textit{70\%}        & \textit{50\%}        \\ \midrule
\multicolumn{7}{c}{\textit{\text{BLEU}}} \\ \midrule
\multicolumn{1}{l|}{\text{CAT DIFF(1,5)}} & 28.3 ± 1.0           & 30.9 ± 1.1           & \multicolumn{1}{r|}{31.0 ± 1.1}          & 34.1 ± 0.7           & 38.1 ± 0.7           & 38.7 ± 0.8           \\
\multicolumn{1}{l|}{\text{Random}}        & \textbf{23.7 ± 1.0}  & \textbf{28.8 ± 1.1}  & \multicolumn{1}{r|}{\text{30.0 ± 1.1}} & \textbf{28.6 ± 0.7}  & \textbf{34.9 ± 0.7}  & \textbf{36.7 ± 0.8}  \\
\multicolumn{1}{l|}{\text{LaBSE}}         & \textbf{5.4 ± 0.6}   & \textbf{23.5 ± 1.0}  & \multicolumn{1}{r|}{\textbf{28.3 ± 1.0}} & \textbf{7.5 ± 0.5}   & \textbf{28.1 ± 0.6}  & \textbf{34.3 ± 0.7}  \\
\multicolumn{1}{l|}{\text{LASER}}         & \textbf{17.1 ± 0.8}  & \textbf{27.8 ± 1.0}  & \multicolumn{1}{r|}{\textbf{30.5 ± 1.1}} & \textbf{20.8 ± 0.6}  & \textbf{33.3 ± 0.7}  & \textbf{36.7 ± 0.8}  \\
\multicolumn{1}{l|}{\text{COMET}}         & \textbf{16.7 ± 0.9}  & \textbf{28.5 ± 1.1}  & \multicolumn{1}{r|}{30.7 ± 1.1}          & \textbf{19.4 ± 0.6}  & \textbf{33.9 ± 0.7}  & \textbf{37.8 ± 0.7}  \\
\multicolumn{1}{l|}{\text{BLOOM 560m}}    & \textbf{23.0 ± 1.0}  & \textbf{27.7 ± 1.0}  & \multicolumn{1}{r|}{\textbf{29.5 ± 1.1}} & \textbf{28.0 ± 0.6}  & \textbf{33.8 ± 0.7}  & \textbf{36.6 ± 0.7}  \\
\multicolumn{1}{l|}{\text{BLOOM 1b}}      & \textbf{22.6 ± 1.0}  & \textbf{27.6 ± 1.0}  & \multicolumn{1}{r|}{\textbf{29.5 ± 1.0}} & \textbf{27.7 ± 0.6}  & \textbf{33.9 ± 0.7}  & \textbf{36.7 ± 0.7}  \\
\multicolumn{1}{l|}{\text{BLOOM 1.7b}}    & \textbf{22.3 ± 0.9}  & \textbf{26.9 ± 1.0}  & \multicolumn{1}{r|}{\textbf{29.9 ± 1.1}} & \textbf{27.1 ± 0.6}  & \textbf{33.5 ± 0.7}  & \textbf{35.9 ± 0.7}  \\
\multicolumn{1}{l|}{\text{Full}}          & 31.5 ± 1.2  & 31.5 ± 1.2  & \multicolumn{1}{r|}{31.5 ± 1.2}          & 38.8 ± 0.8  & 38.8 ± 0.8  & 38.8 ± 0.8           \\ \midrule
\multicolumn{7}{c}{\textit{\text{chrF++}}} \\ \midrule
\multicolumn{1}{l|}{\text{CAT DIFF(1,5)}} & 55.3 ± 0.8           & 57.5 ± 0.7           & \multicolumn{1}{r|}{57.7 ± 0.8}          & 59.2 ± 0.5           & 62.2 ± 0.5           & 62.6 ± 0.5           \\
\multicolumn{1}{l|}{\text{Random}}        & \textbf{51.5 ± 0.7}  & \textbf{55.5 ± 0.8}  & \multicolumn{1}{r|}{\textbf{56.8 ± 0.8}} & \textbf{54.6 ± 0.5}  & \textbf{59.5 ± 0.5}  & \textbf{61.2 ± 0.5}  \\
\multicolumn{1}{l|}{\text{LaBSE}}         & \textbf{24.6 ± 0.7}  & \textbf{51.4 ± 0.8}  & \multicolumn{1}{r|}{\textbf{55.6 ± 0.8}} & \textbf{27.9 ± 0.4}  & \textbf{54.6 ± 0.5}  & \textbf{59.4 ± 0.5}  \\
\multicolumn{1}{l|}{\text{LASER}}         & \textbf{44.5 ± 0.8}  & \textbf{55.3 ± 0.8}  & \multicolumn{1}{r|}{\textbf{57.2 ± 0.8}} & \textbf{47.4 ± 0.5}  & \textbf{58.8 ± 0.5}  & \textbf{61.2 ± 0.5}  \\
\multicolumn{1}{l|}{\text{COMET}}         & \textbf{44.1 ± 0.7}  & \textbf{55.2 ± 0.8}  & \multicolumn{1}{r|}{\textbf{57.1 ± 0.8}} & \textbf{46.2 ± 0.5}  & \textbf{58.8 ± 0.5}  & \textbf{61.6 ± 0.5}  \\
\multicolumn{1}{l|}{\text{BLOOM 560m}}    & \textbf{50.9 ± 0.7}  & \textbf{54.8 ± 0.8}  & \multicolumn{1}{r|}{\textbf{56.2 ± 0.7}} & \textbf{54.3 ± 0.5}  & \textbf{58.8 ± 0.5}  & \textbf{60.9 ± 0.5}  \\
\multicolumn{1}{l|}{\text{BLOOM 1b}}      & \textbf{50.7 ± 0.7}  & \textbf{54.8 ± 0.7}  & \multicolumn{1}{r|}{\textbf{56.2 ± 0.7}} & \textbf{54.0 ± 0.5}  & \textbf{59.0 ± 0.5}  & \textbf{60.9 ± 0.5}  \\
\multicolumn{1}{l|}{\text{BLOOM 1.7b}}    & \textbf{50.3 ± 0.7}  & \textbf{54.5 ± 0.7}  & \multicolumn{1}{r|}{\textbf{56.5 ± 0.8}} & \textbf{53.5 ± 0.5}  & \textbf{58.6 ± 0.5}  & \textbf{60.5 ± 0.5}  \\
\multicolumn{1}{l|}{\text{Full}}          & 57.7 ± 0.8           & 57.7 ± 0.8           & \multicolumn{1}{r|}{57.7 ± 0.8}          & 62.7 ± 0.5  & 62.7 ± 0.5  & 62.7 ± 0.5           \\ \midrule
\multicolumn{7}{c}{\textit{\text{COMET}}} \\ \midrule
\multicolumn{1}{l|}{\text{CAT DIFF(1,5)}} & 77.80                 & 80.38                & \multicolumn{1}{c|}{80.93}               & 78.92                & 82.11                & 82.23                \\
\multicolumn{1}{l|}{\text{Random}}        & \textbf{70.77}       & \textbf{78.70}        & \multicolumn{1}{c|}{\textbf{79.59}}      & \textbf{71.30}        & \textbf{79.14}       & \textbf{80.83}       \\
\multicolumn{1}{l|}{\text{LaBSE}}         & \textbf{46.39}       & \textbf{69.46}       & \multicolumn{1}{c|}{\textbf{76.43}}      & \textbf{48.09}       & \textbf{70.75}       & \textbf{78.00}          \\
\multicolumn{1}{l|}{\text{LASER}}         & \textbf{60.26}       & \textbf{76.14}       & \multicolumn{1}{c|}{\textbf{79.31}}      & \textbf{61.04}       & \textbf{77.08}       & \textbf{80.48}       \\
\multicolumn{1}{l|}{\text{COMET}}         & \textbf{57.70}        & \textbf{78.46}       & \multicolumn{1}{c|}{81.69}               & \textbf{58.65}       & \textbf{78.80}        & 82.59                \\
\multicolumn{1}{l|}{\text{BLOOM 560m}}    & \textbf{69.70}        & \textbf{77.34}       & \multicolumn{1}{c|}{\textbf{79.00}}         & \textbf{71.18}       & \textbf{78.15}       & \textbf{80.93}       \\
\multicolumn{1}{l|}{\text{BLOOM 1b}}      & \textbf{70.35}       & \textbf{76.82}       & \multicolumn{1}{c|}{\textbf{79.25}}      & \textbf{70.72}       & \textbf{78.17}       & \textbf{80.69}       \\
\multicolumn{1}{l|}{\text{BLOOM 1.7b}}    & \textbf{69.58}       & \textbf{76.83}       & \multicolumn{1}{c|}{\textbf{78.38}}      & \textbf{70.14}       & \textbf{77.77}       & \textbf{80.06}       \\
\multicolumn{1}{l|}{\text{Full}}          & 80.86                & 80.86                & \multicolumn{1}{c|}{80.86}               & 82.81                & 82.81                & 82.81    \\
\toprule
\end{tabular}}
\caption{Significance test for German with the null hypothesis: mean translation score of CAT-DIFF is equal to the mean of the other technique. A bolded entry signifies that CAT-DIFF is significantly better (p-value $< 0.05$) than the method used for the corresponding value. For each dataset, we show results for different prune levels (90\%, 70\%, 50\%) and metrics (BLEU, chrF++, COMET). Here "Full" means the value obtained when using all the training data.}
\label{tab:german}
\end{table*}

\begin{table*}[t]
\small
\centering
\resizebox{0.9\textwidth}{!}{
\begin{tabular}{lcccccc}
\toprule
\multicolumn{1}{l}{{\textit{French}}}                       & \multicolumn{3}{c}{\textit{FLORES}}                                                   & \multicolumn{3}{c}{\textit{WMT15}}                                 \\ \cmidrule{2-7}
\multicolumn{1}{l}{\textbf{}}              & \textit{90\%}        & \textit{70\%}        & \multicolumn{1}{c}{\textit{50\%}}       & \textit{90\%}        & \textit{70\%}        & \textit{50\%}        \\ \midrule
\multicolumn{7}{c}{\textit{\text{BLEU}}} \\ \midrule
\multicolumn{1}{l|}{\text{CAT DIFF(1,5)}} & 33.5 ± 1.1           & 40.1 ± 1.3           & \multicolumn{1}{r|}{41.0 ± 1.3}          & 28.2 ± 1.0           & 32.8 ± 1.1           & 33.5 ± 1.1           \\
\multicolumn{1}{l|}{\text{Random}}        & 32.9 ± 1.2           & \textbf{38.6 ± 1.3}  & \multicolumn{1}{r|}{40.9 ± 1.3}          & \textbf{26.9 ± 1.0}  & \textbf{31.0 ± 1.1}  & \textbf{32.5 ± 1.1}  \\
\multicolumn{1}{l|}{\text{LaBSE}}         & \textbf{31.4 ± 1.3}  & 40.1 ± 1.4           & \multicolumn{1}{r|}{41.6 ± 1.4}          & \textbf{25.1 ± 1.1}  & \textbf{31.2 ± 1.1}  & 33.1 ± 1.1           \\
\multicolumn{1}{l|}{\text{Full}}          & 42.4 ± 1.3  & 42.4 ± 1.3  & \multicolumn{1}{r|}{42.4 ± 1.3} & 33.1 ± 1.1  & 33.1 ± 1.1           & 33.1 ± 1.1           \\ \midrule
\multicolumn{7}{c}{\textit{\text{chrF++}}} \\ \midrule
\multicolumn{1}{l|}{\text{CAT DIFF(1,5)}} & 57.5 ± 0.8           & 62.6 ± 0.9           & \multicolumn{1}{r|}{63.0 ± 0.9}          & 53.1 ± 0.8           & 56.7 ± 0.9           & 57.3 ± 0.9           \\
\multicolumn{1}{l|}{\text{Random}}        & 57.3 ± 0.9           & \textbf{61.4 ± 0.9}  & \multicolumn{1}{r|}{62.9 ± 1.0}          & \textbf{51.8 ± 0.8}  & \textbf{55.5 ± 0.8}  & \textbf{56.6 ± 0.8}  \\
\multicolumn{1}{l|}{\text{LaBSE}}         & \textbf{55.3 ± 1.1}  & 62.3 ± 1.0           & \multicolumn{1}{r|}{63.7 ± 1.0}          & \textbf{49.4 ± 1.1}  & \textbf{55.9 ± 0.9}  & 57.4 ± 0.9           \\
\multicolumn{1}{l|}{\text{Full}}          & 63.9 ± 0.9  & 63.9 ± 0.9  & \multicolumn{1}{r|}{63.9 ± 0.9}          & 57.3 ± 0.9  & 57.3 ± 0.9  & 57.3 ± 0.9           \\ \midrule
\multicolumn{7}{c}{\textit{\text{COMET}}} \\ \midrule
\multicolumn{1}{l|}{\text{CAT DIFF(1,5)}} & 77.15                & 82.60                 & \multicolumn{1}{c|}{82.94}               & 72.50                 & 77.19                & 78.08                \\
\multicolumn{1}{l|}{\text{Random}}        & \textbf{75.44}       & \textbf{81.02}       & \multicolumn{1}{c|}{\textbf{82.41}}      & \textbf{70.74}       & \textbf{75.58}       & \textbf{77.02}       \\
\multicolumn{1}{l|}{\text{LaBSE}}         & \textbf{73.81}       & \textbf{81.22}       & \multicolumn{1}{c|}{82.86}               & \textbf{69.72}       & \textbf{75.82}       & \textbf{77.28}       \\
\multicolumn{1}{l|}{\text{Full}}          & 83.78                & 83.78                & \multicolumn{1}{c|}{83.78}               & 78.31                & 78.31                & 78.31      \\
\bottomrule
\end{tabular}}
\caption{Significance test for French with the null hypothesis: mean translation score of CAT-DIFF is equal to the mean of the other technique. A bolded entry signifies that CAT-DIFF is significantly better (p-value $< 0.05$) than the method used for the corresponding value. For each dataset, we show results for different prune levels (90\%, 70\%, 50\%) and metrics (BLEU, chrF++, COMET). Here "Full" means the value obtained when using all the training data.
}
\label{tab:french}
\end{table*}


\end{document}